\documentclass[runningheads]{llncs}
\usepackage{graphicx}
\usepackage{subcaption}
\usepackage{wrapfig}
\usepackage{xcolor}
\usepackage{array}
\usepackage{float}
\usepackage{theorem}
\usepackage{textfit}
\usepackage{placeins}
\usepackage{framed, color}
\usepackage{url}
\usepackage{rotating}
\usepackage{times}
\usepackage{amsmath}
\usepackage{amsfonts}
\usepackage{algpseudocode}
\usepackage{algorithm}
\usepackage{tabularx}
\usepackage{booktabs}
\usepackage{ragged2e}
\usepackage{listings}
\usepackage[breaklinks]{hyperref}
\usepackage{tabto}
\usepackage{lipsum}
\usepackage[export]{adjustbox}
\usepackage{caption}
\usepackage{multirow}
\usepackage{enumitem}
\usepackage{xcolor}
\usepackage[justification=centering]{caption}

\usepackage{listings}
\begin{document}
\title{Survival Meets Classification: A Novel Framework for Early Risk Prediction Models of Chronic Diseases}
\titlerunning{Chronic Diseases Survival Classification}
%
\author{Shaheer Ahmad Khan\inst{1}\orcidID{0009-0000-8772-8149}  \and Muhammad Usamah Shahid\inst{1}\orcidID{0009-0001-4293-2979}  \and
Muddassar Farooq\inst{1}}
\authorrunning{S. A. Khan, U. Shahid, M. Farooq}
%
\institute{CureMD Research, 80 Pine St 21st Floor, New York, NY 10005, United States
\email{\{shaheer.ahmed, muhammad.usamah, muddassar.farooq\}@curemd.com}\newline
\url{https://www.curemd.com/}}
\maketitle              
\begin{abstract}
Chronic diseases are long-lasting conditions that require lifelong medical attention. Using big EMR data, we have developed early disease risk prediction models for five common chronic diseases: diabetes, hypertension, CKD, COPD, and chronic ischemic heart disease. In this study, we present a novel approach for disease risk models by integrating survival analysis with classification techniques. Traditional models for predicting the risk of chronic diseases predominantly focus on either survival analysis or classification independently. In this paper, we show survival analysis methods can be re-engineered to enable them to do classification efficiently and effectively, thereby making them a comprehensive tool for developing disease risk surveillance models. The results of our experiments on real-world big EMR data show that the performance of survival models in terms of accuracy, F1 score, and AUROC is comparable to or better than that of prior state-of-the-art models like LightGBM and XGBoost. Lastly, the proposed survival models use a novel methodology to generate explanations, which have been clinically validated by a panel of three expert physicians.
\end{abstract}
\section{Introduction}

Chronic diseases represent persistent health conditions that require sustained patient management. These diseases not only severely impact the routine activities of patients but also are a leading cause of death and disability worldwide. Moreover, the management of these diseases is also significantly escalating healthcare budgets. Conditions such as hypertension and diabetes are particularly insidious, often leading to a set of complications and severe morbidity. To this end, our objective is to design an early-warning disease surveillance system capable of issuing timely alerts. Such alerts would enable timely medical, lifestyle, diet, or other preventive interventions to mitigate the onset and progression of chronic diseases. Our partner physicians put a challenging design requirement for our early warning system to make it relevant to the real world: it must only use regular patients' data routinely recorded in Electronic Medical Records (EMR) systems excluding the labs; as a result, its usefulness will be significantly enhanced as it could issue early alerts well before the time when a healthcare provider begins to suspect the onset of a condition and starts lab investigations.

The application of Machine Learning (ML) in disease prediction has received increasing attention within the scientific community. Key focal areas for such models include diabetes\cite{dinh2019data,zou2018predicting}, hypertension\cite{martinez2021review}, cardiovascular\cite{dinh2019data}, and kidney diseases\cite{ilyas2021chronic}. A comprehensive review covering ML techniques developed between 2012 and 2021 was conducted by the authors of \cite{ahsan2022machine}. However, it is important to note that the majority of these studies focus on developing disease classifiers that determine the current diagnosis of a patient and do not predict future disease onset.

The body of research dedicated to early disease prediction remains relatively less explored. Studies have independently employed both classification and survival analysis models to tackle this challenge. For instance, a recent study utilized gradient boosting to forecast the occurrence of end-stage kidney disease within a two-year time frame \cite{petousis2024early}. In the domain of Chronic Kidney Disease (CKD), another investigation applied XGBoost, albeit relying on a constrained set of predictors, to predict early diagnosis \cite{islam2023chronic}. Similarly, the risk of developing Type 2 Diabetes within the next year was predicted using random forests in a related study\cite{mani2012type}.

A common feature of these research efforts is their use of laboratory test results for predictive features, including HbA1c, serum creatinine, eGFR, lipid profiles, BUN, among others. This approach, while valuable in the context of determining the current diagnosis of a patient may not be relevant to our work, as our aim is to predict early without lab results at a time when healthcare providers typically are not suspecting the risk. Moreover, classification models also do not allow physicians to observe a disease risk progression over time, a valuable feature for planning timely interventions. Survival models address these challenges by offering a continuous risk assessment over time and avoiding the need for multiple independent classifiers for different time periods, thus providing an integrated and clinically relevant solution for early disease risk prediction.

Several studies have utilized survival models for predicting the onset or diagnosis of diseases such as Diabetes\cite{liu2018early}, Chronic Kidney Disease (CKD)\cite{hagar2014survival}, and Hypertension\cite{migora2021survival}. Additionally, a comprehensive review focusing on Chronic Obstructive Pulmonary Disease (COPD) and survival analysis has been conducted in \cite{matheson2018prediction}. Survival studies typically gauge their effectiveness using the concordance index (C-index), complicating direct comparisons with classifiers designed for similar use cases. The literature provides limited guidance on transforming survival model outputs into classification predictions or assessing them using classification metrics. While one study reports promising classification metrics, it falls short of explaining the methodology for adapting survival models for doing classification \cite{zhou2022development}. Another study suggests the adjustment of survival probability thresholds over time to enable classification \cite{weiser2011predicting}. A similar study, confronting similar issues, opts to train separate classifiers within a survival framework \cite{ahmad2023classification}. Similarly, an XGBoost classifier and a multivariate Cox regression analysis are independently employed by one study to predict the incidence of hypertension \cite{ye2018prediction}. Our study aims to bridge this gap by offering a strategy in which survival models are re-engineered to also act as effective classifiers.
\begin{table}[ht]
\vspace{-0.5cm}
	\caption{Patient counts for each chronic disease that were eligible for model training, and their respective criteria}
	\label{tab : data_overview}
	\centering
	\begin{tabular}{c|c| c }
		\hline
		Disease & ICD10 codes &  Patient Count \\
		\hline
		Hypertension (HTN) & I10-I13 & 54654 \\
		Type 2 Diabetes Mellitus (DM) & E11 &  31865 \\
		Chronic Kidney Disease (CKD) & N18, I12, I13, [E08-E13].22 & 15316\\
		Chronic Ischemic Heart Disease (CHD) & I25 &  13804 \\
		Chronic Obstructive Pulmonary Disease (COPD) & J41-J44 & 15131 \\
		\hline
	\end{tabular}

\end{table}

The major contributions of this paper are: (1) re-engineering survival models such that they could be used to derive classification inferences from survival analysis; (2) creating novel early disease risk prediction models, excluding the lab reports that confirm the diagnosis of the chronic disease, to assist physicians in preventive management of patients including lifestyle changes and/or diet; (3) using a novel method to explain the survival models with the help of SHAP algorithm; (4) generalizing the disease risk prediction models to five most prevalent chronic diseases - some of which are underrepresented in current literature; and (5) validating features' set, risk factors, model engineering workflow, and explanations by a panel of three expert physicians making the study outcomes clinically relevant and rooted in sound medical knowledge.

\section{Data Overview}
Patient Electronic Medical Records (EMRs) from various client practices are de-identified and anonymized in compliance with HIPAA guidelines, and subsequently aggregated into an in-house data lakehouse. The data pre-processing phase encompassed noise elimination (such as duplicates, errors, missing values, or test entries), standardization, mappings, and specific cohort selection for each chronic disease under study.

EMR data is inherently temporal, accruing incremental information with each patient interaction within the healthcare delivery system. To qualify for inclusion in our study, patients must have had a sufficient number of encounters before the time when a chronic disease is diagnosed. Table \ref{tab : data_overview} tabulates the patient count eligible for model training, ensuring a robust dataset for our analysis. The earliest occurrence of any of the diagnosis codes listed in the table is considered the diagnosis date.

For normal datasets, we ensured that patients had at least three encounters spanning at least one year, in line with our objective of predicting the 12-month risk of chronic diseases. We applied respective exclusion criteria on approximately 10 million patients and performed significant random under-sampling to create balanced datasets. 

\textbf{Feature Engineering.} A panel of three physicians at a partner clinic was instrumental in designing and selecting features that deemed risk factors for or indicative of the chosen chronic diseases. These features comprise patient demographics (age, race, gender), diagnosis records (using ICD-10 codes, Elixhauser comorbidity groups, or custom groupings), vital signs, medications (categorized by therapeutic properties via GPI codes), as well as social and family history. To comply with our analytical framework, we treated all features as categorical: continuous values were grouped into bins, and discrete, non-numeric information was systematically encoded. A complete list of features used in our five chronic disease models is provided in Appendix \ref{ap: features}.

\section{Methodology}
\subsection{Data preparation approaches}\label{sec: data_prep}
In our retrospective study, leveraging historical data gathered over an extended period, the application of survival analysis poses distinct challenges. A straightforward method involves selecting a fixed time period and analyzing the data within this time window as if it were gathered explicitly for a survival study. This method, however, might lead to significant data loss, particularly when dealing with narrow time frames (e.g., a 1-year window in our study). To preserve information, we applied our chosen time windows to each patient's data.

For robust early prediction modeling, it's imperative to identify a cutoff point prior to disease diagnosis (or the latest encounter for patients without the disease), beyond which a patient's recorded data is excluded from the analysis. For patients diagnosed with a disease, we determine this cutoff by asking: "At which encounter would it have been beneficial for my model to raise an alert?" Adhering to the conventions of survival analysis, we consistently selected the patient's earliest encounter within our predetermined window which is the year leading up to the diagnosis.

For patients without a diagnosis, we explored three distinct methodologies: (1) Mirroring the method for diseased patients, we considered the earliest encounter within the year leading up to their last recorded encounter, and this aligns with traditional survival studies, where the observation time is maximally one year; (2) Departing from the 1-year constraint, opting instead for the patient's second encounter, regardless of its timing relative to our specified window; and (3) Selecting the latest encounter occurring before the start of the designated 1-year window, as this strategy is closely aligned with classification approaches and eliminates any time overlap between the datasets of event and non-event cases (It aims to utilize the encounter offering the maximum available patient information while ensuring no diagnosis was made in the year following that encounter).

\begin{figure}
\vspace{-0.5cm}
	\centering
	\begin{subfigure}[b]{0.32\textwidth}
		\includegraphics[width=\textwidth]{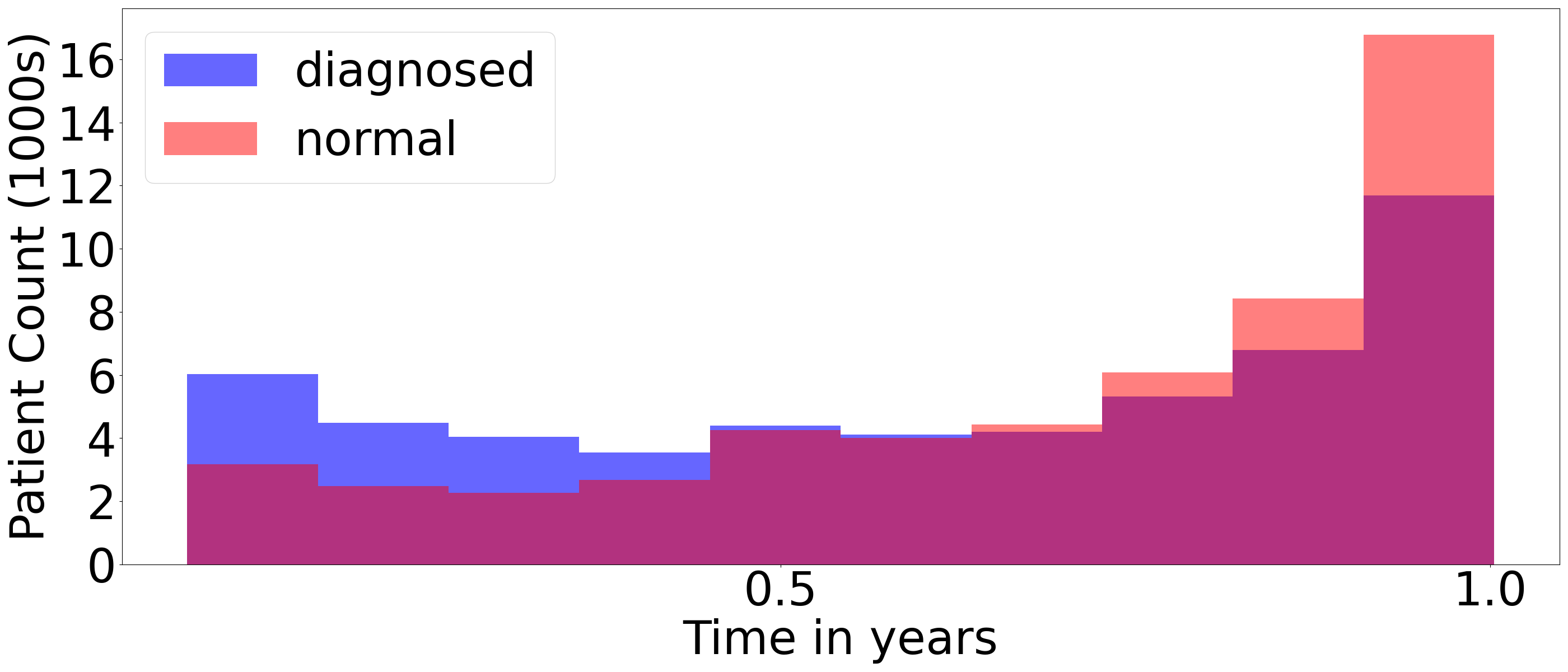}
		\caption{Approach 1: Similar} \label{fig : surv_1}
	\end{subfigure}
\begin{subfigure}[b]{0.32\textwidth}
	\includegraphics[width=\textwidth]{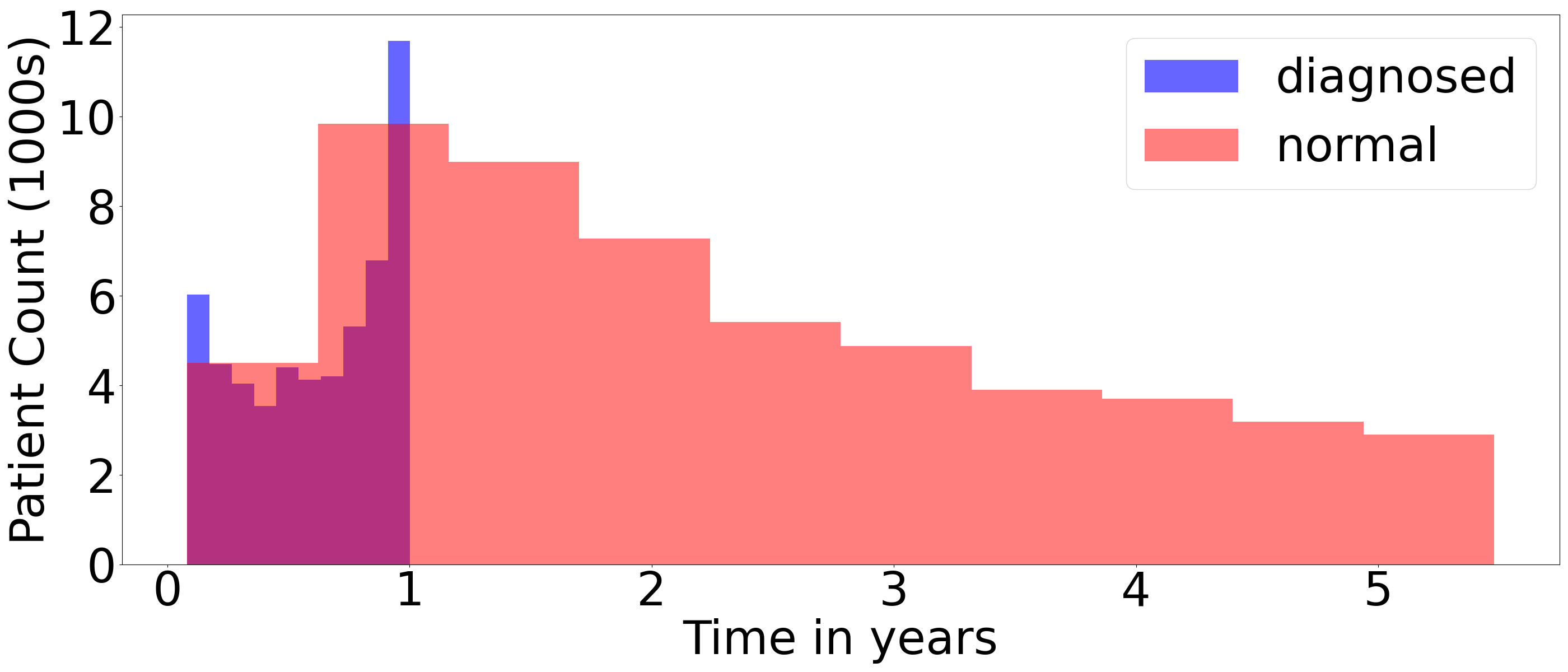}
	\caption{Approach 2: Overlap} \label{fig : mean_mad}
\end{subfigure}
\begin{subfigure}[b]{0.32\textwidth}
	\includegraphics[width=\textwidth]{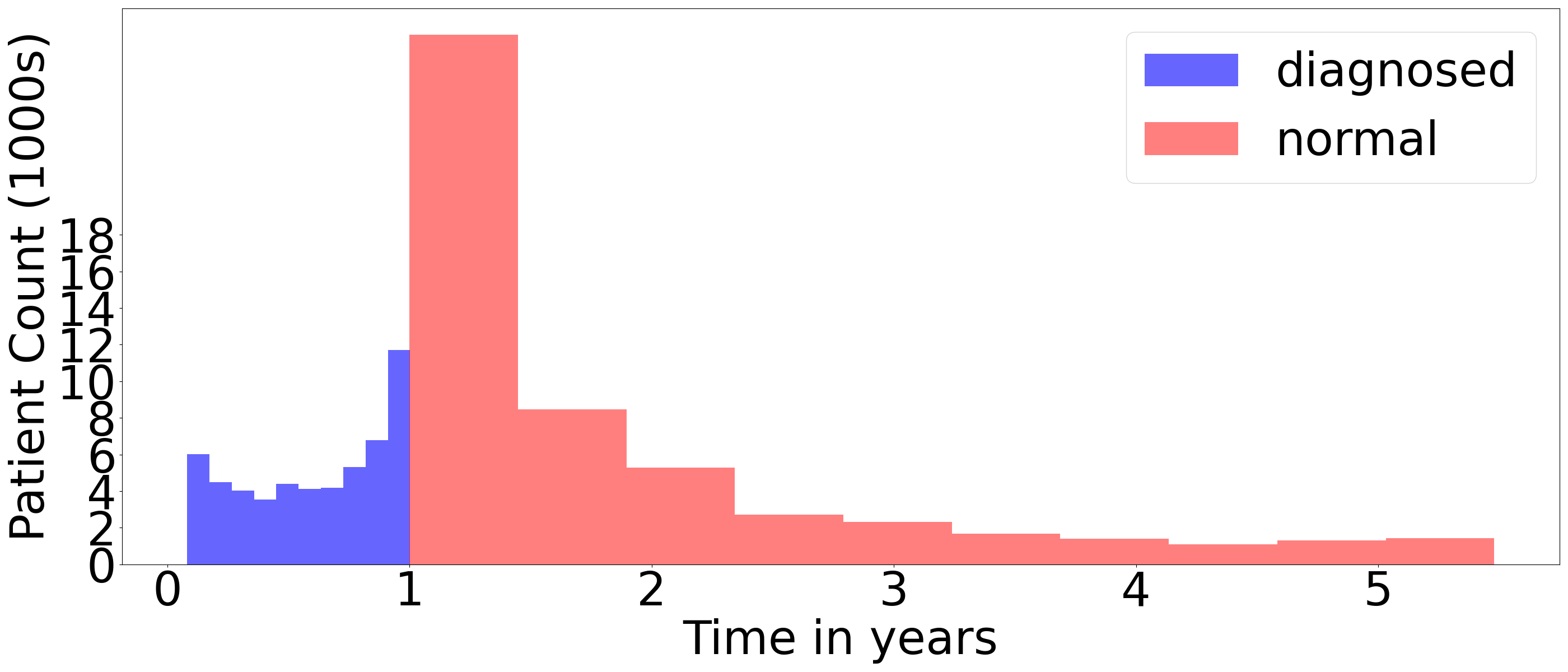}
	\caption{Approach 3: Distinct} \label{fig : mean_mad}
\end{subfigure}
	\caption{Distribution of observation times for hypertension using the three described approaches. Similar plots for the remaining diseases can be found in Appendix \ref{ap: time_distributions}.}\label{fig: time distribution}
\vspace{-0.5cm}
\end{figure}

These approaches are visualized in Figure \ref{fig: time distribution}. The descriptive names, detailed in the figure captions, will be used to refer to these approaches throughout the remainder of this document. We discuss our motivation behind exploring alternative approaches in Section \ref{sec: results}.

\subsection{Classification from survival models}\label{sec: cls_from_surv}
We explore three techniques in our study to derive classification inferences from survival models. 

\textbf{1. Risk-score based classification (RS).} 
This approach utilizes the survival model's predicted risk score for classification by following three steps: (1) calculate the risk score for each patient in the training set using the trained survival model; (2) determine an optimal threshold for the risk score that maximizes classification accuracy or other desired metrics; and (3) classify each patient's outcome as 1 (disease diagnosed) if its risk score exceeds this threshold, otherwise 0 (no disease). In our implementation, we traversed all predicted risk scores on the training set to identify an effective classification threshold.

\textbf{2. Survival probability at last time step (SP).} 
This method directly applies survival analysis models' key output, survival curves, which illustrate the event's (in this case, disease onset) non-occurrence probability by a certain time. Its three steps are: (1) estimate the survival function for a given patient using the trained model; (2) examine the survival probability at the last time-step, typically the study duration of 1 year; and (3) apply a 0.5 threshold to this probability, akin to traditional classifiers, to determine the disease probability ($P<=0.5$ indicates disease diagnosis).
While studies have sought optimized thresholds for survival probability, we propose that a fixed 0.5 threshold enables a smoother transition to classification inferences from survival models.

\textbf{3. Leaf node analysis (LN).} 
Applicable to tree-based survival models like random survival forests (RSF), this method derives inference by examining the label distribution at a leaf node, similar to classification trees. The three steps are: (1) ensure the necessary information is stored at the leaf nodes of a survival tree, and this will vary in implementation, but a commonly applicable solution is to maintain a separate log of which training samples are at each leaf node, or simply, the ratio of samples by event label; (2) traverse the tree, for a given patient, to identify the corresponding leaf node; and (3) classify the patient based on the known label distribution for this leaf node, and subsequently a probabilistic prediction is made by dividing the number of patients with the disease (label 1) by those without the disease (label 0). For ensemble models like survival forests, predictions from constituent trees can be aggregated as either a majority vote or an average probability.

\section{Results and Discussions}\label{sec: results}

We prepared our feature matrices following the methodologies outlined in the previous sections, employing a 70-10-20 split for training, validation, and testing datasets respectively. We then trained three prominent tree ensemble classifiers -- Random Forests, XGBoost, and LightGBM -- for a comparative analysis. In addition, we also evaluated a Random Survival Forest's performance as a classifier, employing the three techniques described in Section \ref{sec: cls_from_surv}. Performance metrics, including the F1 scores for the ensemble classifiers on the validation set, the concordance index (C-index) for the survival forest, and the F1 scores obtained from the survival forest using the described classification techniques, are tabulated in Table \ref{table: both_within_results}. For consistency and to ensure repeatability, all models were trained using their respective Python libraries' default hyper-parameters.
 \setlength{\tabcolsep}{4pt}
        \renewcommand{\arraystretch}{1.25}
        \begin{table}[ht]
        \vspace{-0.5cm}
            \centering
            \caption{Evaluation of Approach 1: Similar on the validation set.}
            \label{table: both_within_results}
            \begin{tabular}{l|c|c|c|c|c|c|c}
            \hline
             \multicolumn{1}{c|}{} & \multicolumn{3}{c|}{Classifiers} & \multicolumn{4}{c}{Random survival forest} \\

            \hline
            Disease & Random Forest & XGB & LGBM & C-Index & F1 (RS) & F1 (LN) & F1 (SP) \\
            \hline
            Hypertension & 0.728 & 0.751 & 0.751 & 0.689 & 0.762 & 0.749 & 0.743 \\
  
            Heart        & 0.730 & 0.738 & 0.753 & 0.667 & 0.750 & 0.750 & 0.742 \\
    
            CKD          & 0.726 & 0.735 & 0.748 & 0.658 & 0.749 & 0.750 & 0.742 \\

            COPD         & 0.726 & 0.737 & 0.743 & 0.675 & 0.737 & 0.750 & 0.721 \\

            Diabetes     & 0.741 & 0.752 & 0.752 & 0.688 & 0.766 & 0.749 & 0.735 \\
            \hline
            \end{tabular}
            
        \end{table}

\begin{figure}[ht]

	\centering
	\begin{subfigure}[b]{0.32\textwidth}
		\includegraphics[width=\textwidth]{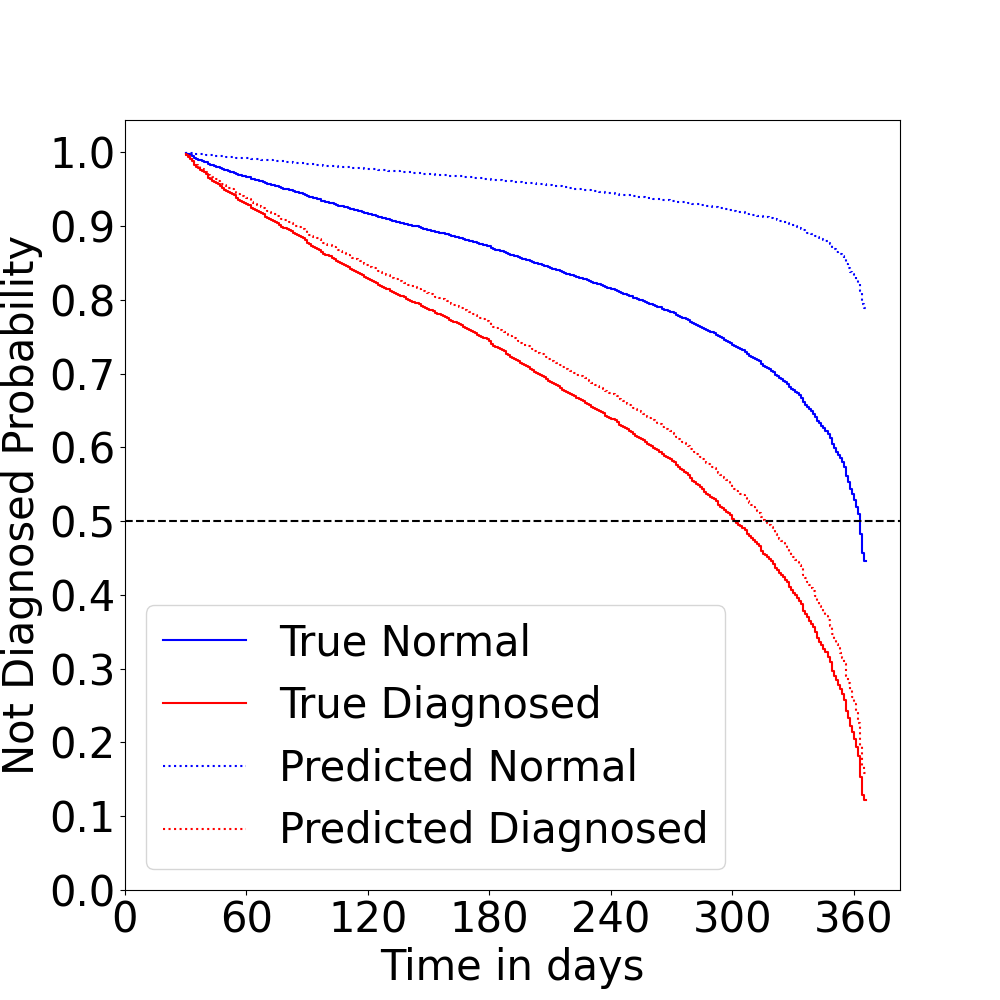}
		\caption{Approach 1} \label{fig : surv_1}
	\end{subfigure}
\begin{subfigure}[b]{0.32\textwidth}
	\includegraphics[width=\textwidth]{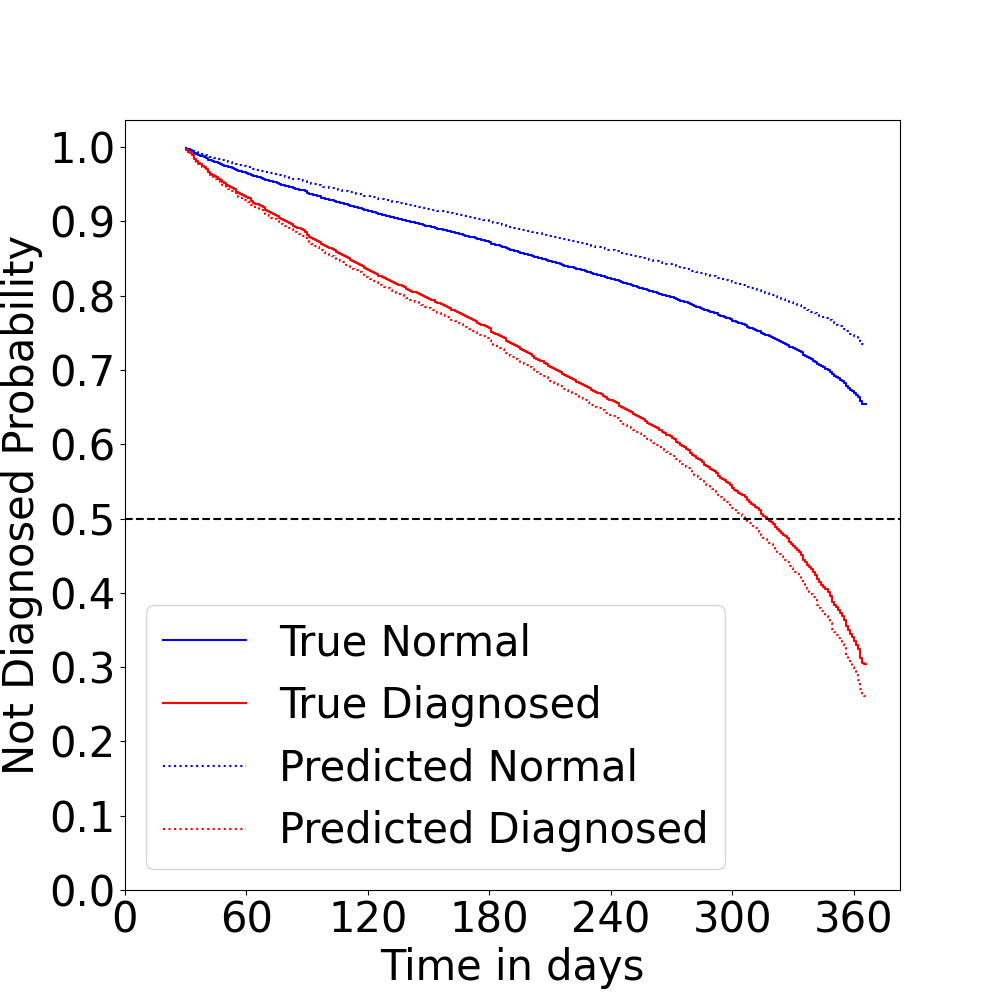}
	\caption{Approach 2} \label{fig : surv_2}
\end{subfigure}
\begin{subfigure}[b]{0.32\textwidth}
	\includegraphics[width=\textwidth]{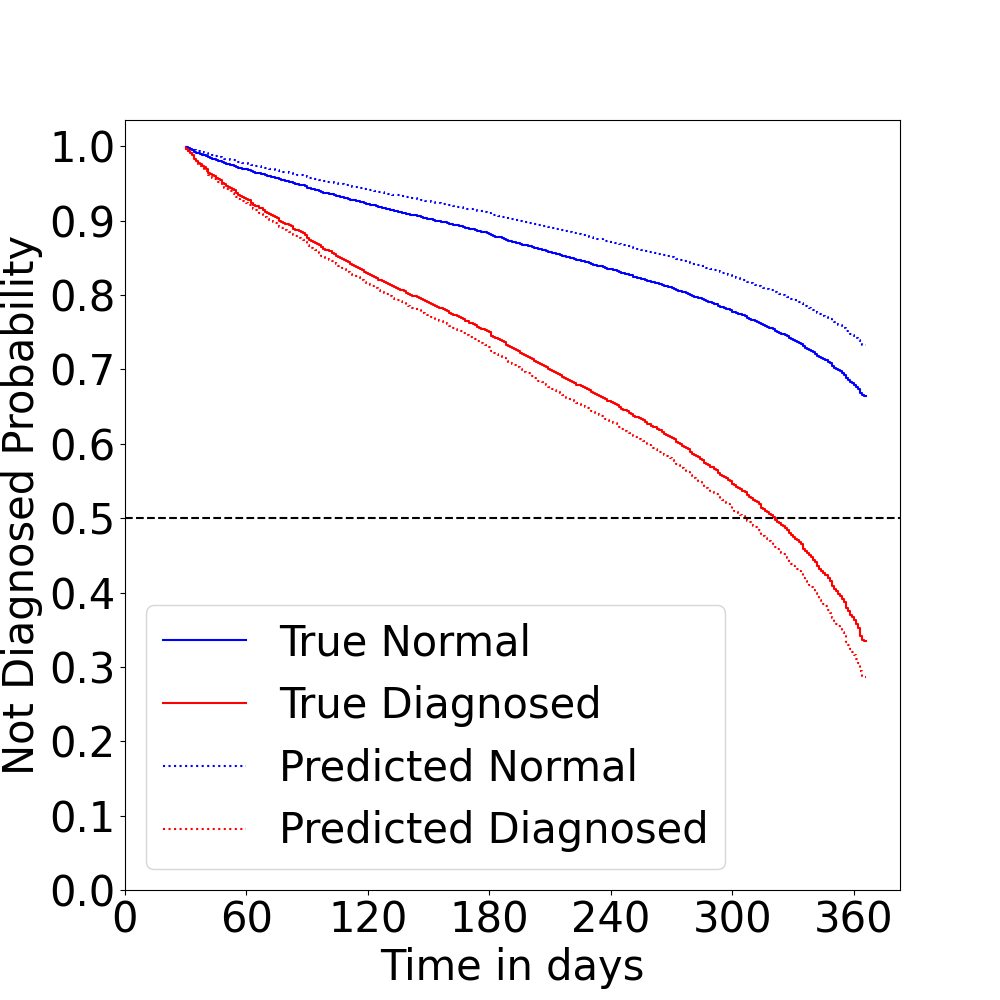}
	\caption{Approach 3} \label{fig : surv_3}
\end{subfigure}
	\caption{Average survival curves for the hypertension training set. Clusters are made using the true and predicted labels found using survival probability for classification. Similar curves for the remaining diseases can be found in Appendix \ref{ap: survival_curves}.}

\end{figure}
Although the survival model's F1 scores are on par with the top-performing classifier, they fell short of being competitive, and the model's concordance index (C-index) is also relatively low. Further investigation into the models' behavior through survival curves reveals a consistent pattern: a marked decline in survival probabilities as the one-year mark approached. This trend, depicted in Figure \ref{fig : surv_1}, not only potentially undermines the model's performance -- evidenced by the average predicted survival probability for true normal patients in hypertension training set falling below the 0.5 threshold -- but also presents a clinical paradox by suggesting a significant increase in patient's risk in the final month, challenging the model's utility for clinical decision-making. We hypothesized that this anomaly arose from the model's lack of exposure to patients surviving past this point; therefore, we developed alternative data preparation methods as discussed in Section \ref{sec: data_prep}.
 \setlength{\tabcolsep}{4pt}
        \renewcommand{\arraystretch}{1.25}
        \begin{table}[!t]
        \vspace{-0.5cm}
            \centering
            \caption{Evaluation of Approach 2: Overlap on the validation set.}
            \label{table: mixed_results}
            \begin{tabular}{l|c|c|c|c|c|c|c}
            \hline
            \multicolumn{1}{c|}{} & \multicolumn{3}{c|}{Classifiers} & \multicolumn{4}{c}{Random survival forest} \\
            \hline
           Disease & Random Forest & XGB & LGBM & C-Index & F1 (RS) & F1 (LN) & F1 (SP) \\
            \hline
            Hypertension &  0.728 & 0.744 & 0.744 & 0.708 & 0.762 & 0.747 & 0.750  \\

            Heart        &  0.799 & 0.808 & 0.814 & 0.720 & 0.813 & 0.815 & 0.796  \\

            CKD          &  0.768 & 0.787 & 0.787 & 0.718 & 0.790 & 0.792 & 0.796   \\
 
            COPD         &  0.788 & 0.785 & 0.790 & 0.731 & 0.794 & 0.798 & 0.787  \\

            Diabetes     &  0.759 & 0.773 & 0.767 & 0.729 & 0.780 & 0.768 & 0.774   \\
            \hline
            \end{tabular}

        \end{table}
\setlength{\tabcolsep}{4pt}
        \renewcommand{\arraystretch}{1.25}
        \begin{table}[!h]
        
            \centering
            \caption{Evaluation of Approach 3: Distinct on the validation set.}
            \label{table: diag_within_normal_without_results}
            \begin{tabular}{l|c|c|c|c|c|c|c}
            \hline
            \multicolumn{1}{c|}{} & \multicolumn{3}{c|}{Classifiers} & \multicolumn{4}{c}{Random survival forest} \\
            \hline
           Disease & Random Forest & XGB & LGBM & C-Index & F1 (RS) & F1 (LN) & F1 (SP) \\
            \hline
            Hypertension &  0.712 & 0.741 & 0.743 & 0.720 & 0.747 & 0.741 & 0.742 \\
 
            Heart        &  0.763 & 0.766 & 0.784 & 0.716 & 0.788 & 0.786 & 0.772  \\

            CKD          &  0.757 & 0.763 & 0.775 & 0.718 & 0.772 & 0.775 & 0.777  \\

            COPD         &  0.769 & 0.772 & 0.773 & 0.729 & 0.775 & 0.778 & 0.761   \\

            Diabetes     &  0.757 & 0.767 & 0.770 & 0.745 & 0.780 & 0.770 & 0.775  \\
            \hline
            \end{tabular}
   
        \end{table}
        
The performance enhancements on the validation sets from these approaches are evident in Tables \ref{table: mixed_results} and \ref{table: diag_within_normal_without_results}. These results underscore LightGBM's superiority as a classifier and reveal a notable improvement in the performance of classifiers and the survival model. However, the comparison between the survival model's effectiveness and the optimal classification technique is still a challenge.

\begin{wrapfigure}{r}{7cm}
\vspace{-0.5cm}
\centering
	\includegraphics[width=7cm]{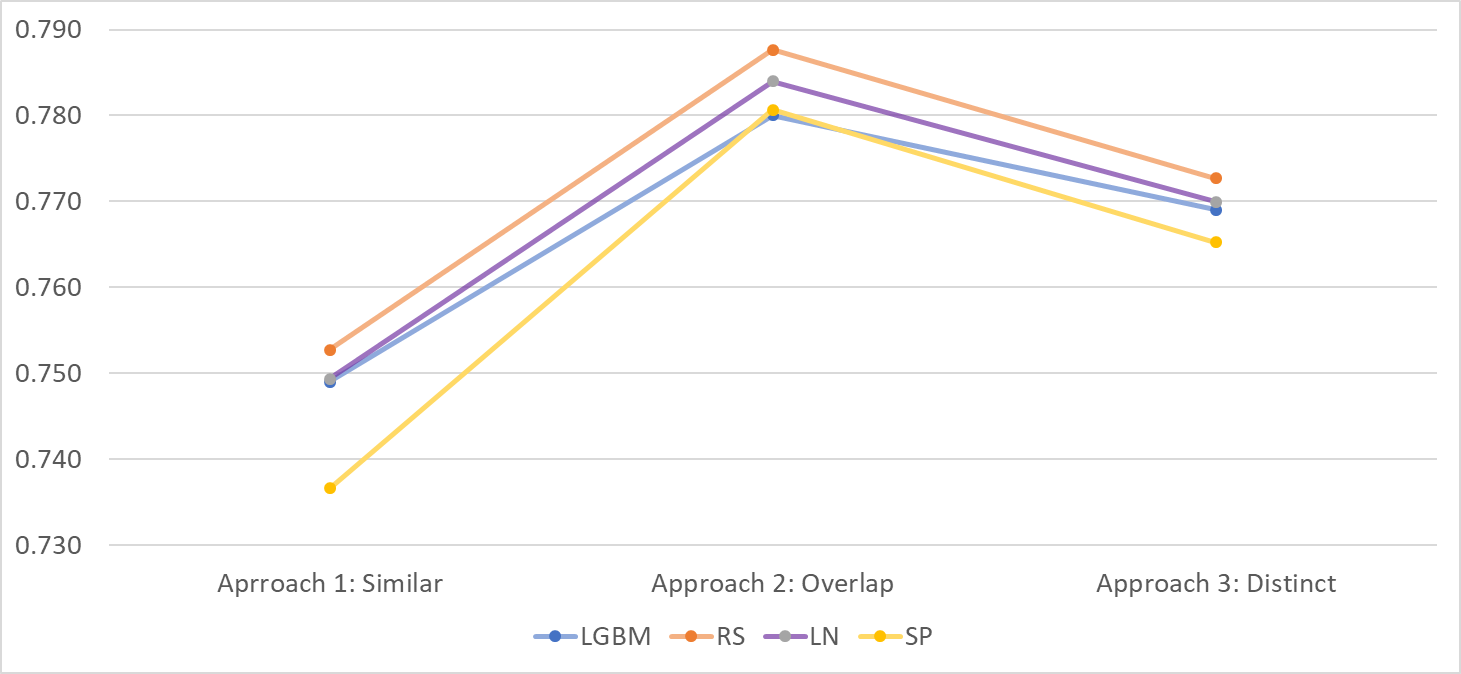}
	\caption{Average F1 scores of LGBM and RSF using the three classification techniques} \label{fig : ave_f1}
 \vspace{-0.5cm}
\end{wrapfigure}

Figure \ref{fig : ave_f1} aims to clarify these comparisons by visualizing the classification metrics outcomes. It shows that our survival forest outperforms the traditional classifiers across all data preparation and classification methodologies. Among the evaluated factors, the choice of dataset preparation method significantly influences the performance, and in comparison, the various classification techniques provide approximately the same performance. Consistently, employing a risk score threshold emerges as the most effective strategy. 
However, due to its practical applicability, the survival probability approach is recommended for real-world implementations. Detailed results utilizing this method on the test set are tabulated in Table \ref{tab: final_res}.

Encouragingly, both the Area Under the Receiver Operating Characteristic (AUROC) and the Area Under the Precision-Recall Curve (AUPRC) metrics are notably high. Within the range of diseases analyzed, hypertension emerges as the most challenging to predict across all metrics.
\begin{table}[ht]
\vspace{-0.5cm}
\centering
\caption{Performance of the Random Survival Forest on the test set.}\label{tab: final_res}
 \resizebox{\columnwidth}{!}{%
\begin{tabular}{l |c| c| c| c| c| c| c| c| c}
\hline
Disease & C Index & Accuracy & Precision & Recall & NPV & Specificity & AUROC & AUPRC & F1 score \\
\hline
Hypertension & 0.709 & 0.742 & 0.723 & 0.779 & 0.764 & 0.705 & 0.828 & 0.819 & 0.755 \\
Heart & 0.741 & 0.788 & 0.758 & 0.838 & 0.823 & 0.739 & 0.869 & 0.852 & 0.819 \\
CKD & 0.729 & 0.789 & 0.767 & 0.827 & 0.814 & 0.751 & 0.870 & 0.859 & 0.796 \\
COPD & 0.730 & 0.784 & 0.761 & 0.815 & 0.809 & 0.753 & 0.869 & 0.871 & 0.799 \\
Diabetes & 0.728 & 0.784 & 0.819 & 0.733 & 0.756 & 0.836 & 0.872 & 0.896 & 0.778 \\
\hline
\end{tabular}
}
\vspace{-0.5cm}
\end{table}

\section{Explainability}

The interpretability of machine learning models, particularly in the healthcare domain, is an important requirement. Clinicians' trust in these models often hinges on their ability to understand how decisions are made by them. In survival analysis, Cox Regression models are favored for their interpretability since the regression coefficients of various features are explicitly known. However, in many research contexts, the performance of Cox Regression models is surpassed by Random Survival Forests (RSF), which are more complex and considered black box models. 

In the literature, we identified a few notable attempts to explain RSFs. SurvSHAP \cite{survSHAP} operates as a surrogate-based algorithm. It initially identifies the optimal number of survival function clusters through log-rank difference analysis, then trains multiple Random Forest Classifiers (equal to the number of optimal clusters) to predict each cluster. The Shap's TreeExplainer is subsequently used to explain the decision-making of these classifiers, by aggregating the importance of features across different clusters for comprehensive insight. Several other methods employ similar surrogate models for explanations.

\begin{figure}[ht]
\vspace{-0.5cm}
	\centering
	\begin{subfigure}[b]{0.4\textwidth}
		\includegraphics[width=\textwidth]{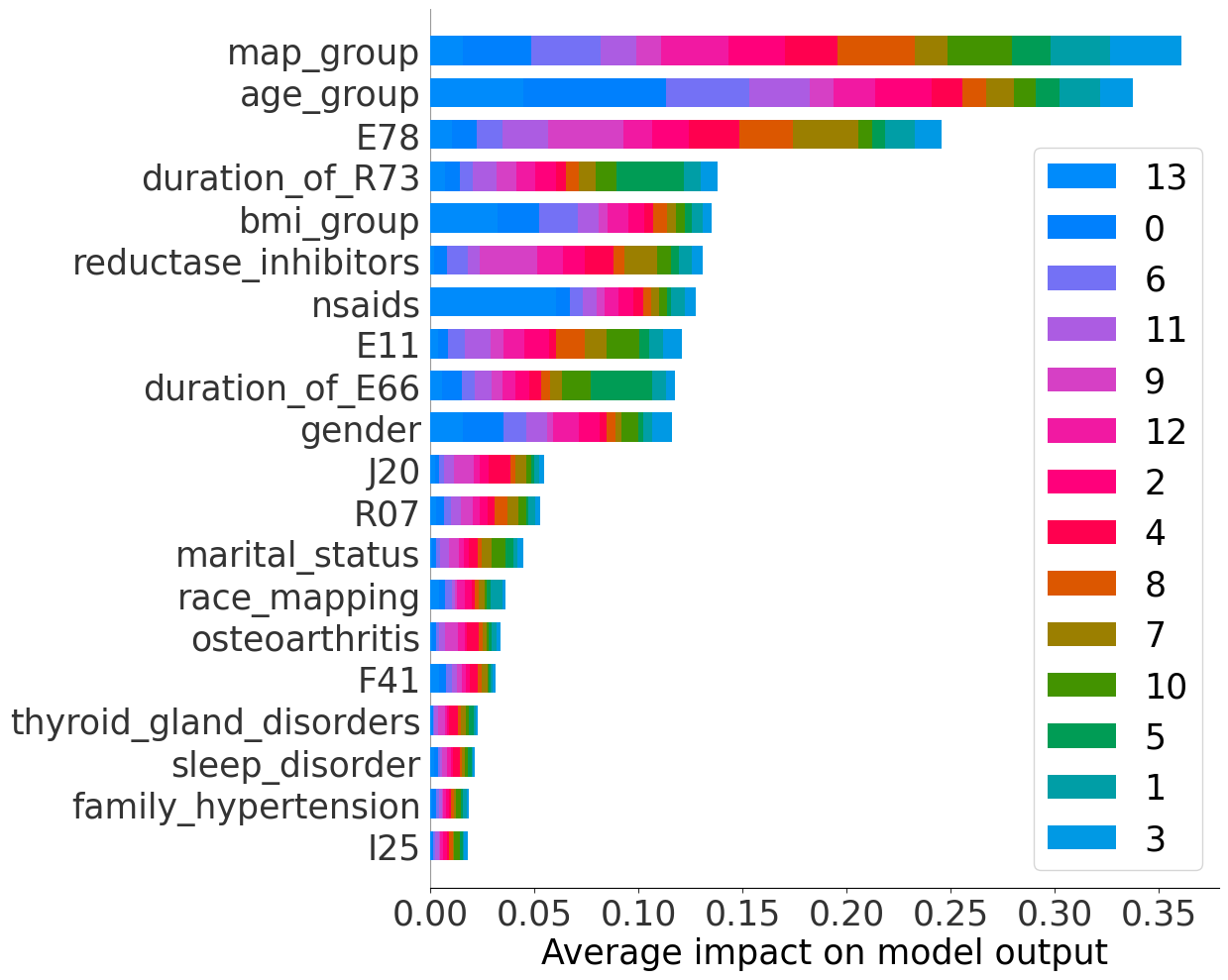}
		\caption{SurvSHAP} \label{fig : htn_survshap}
	\end{subfigure}
\begin{subfigure}[b]{0.4\textwidth}
	\includegraphics[width=\textwidth]{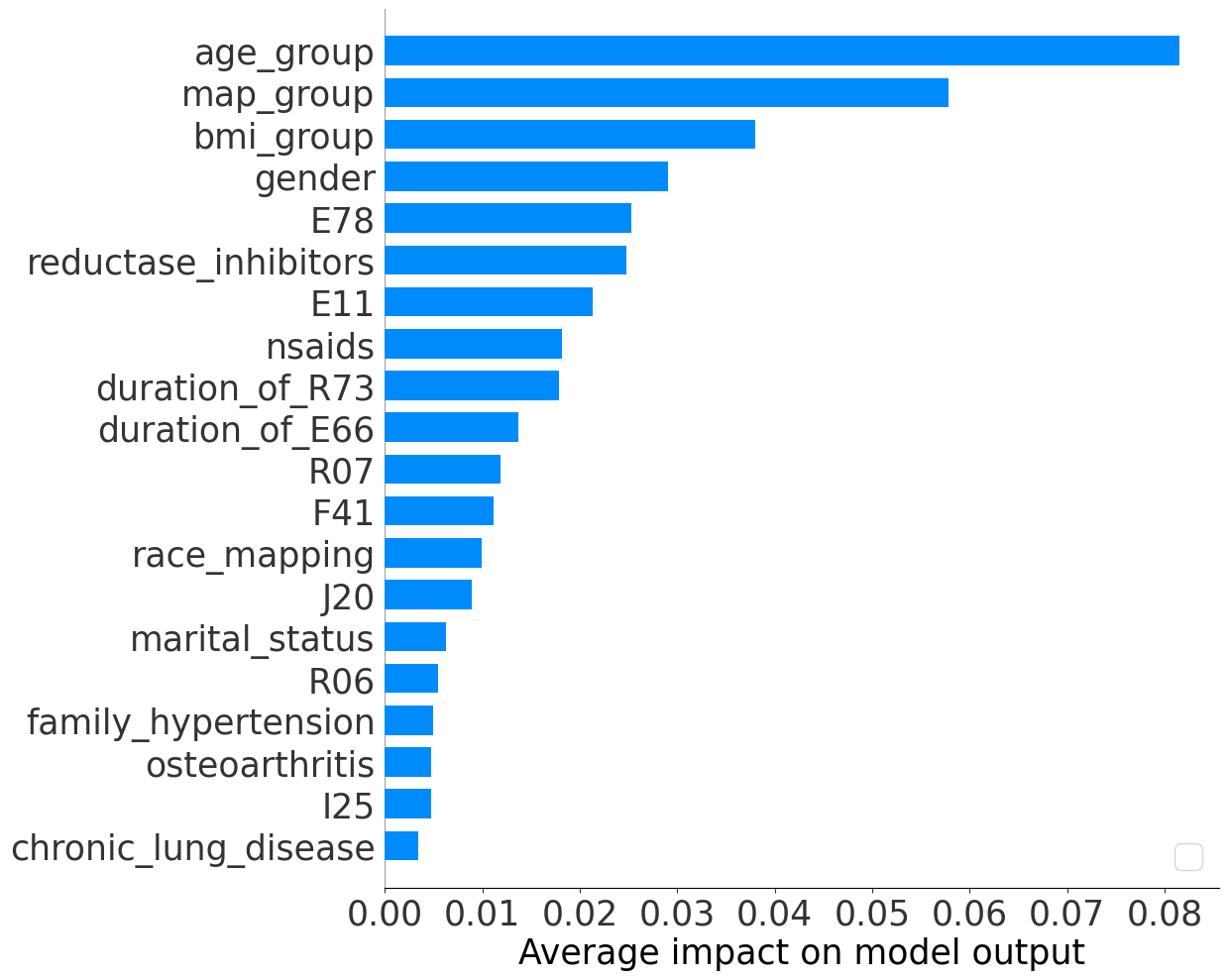}
	\caption{Custom Implementation} \label{fig : htn_selfsurv}
\end{subfigure}
	\caption{Feature importances found using SurvSHAP and our custom implementation. Similar plots for the remaining diseases can be found in Appendix \ref{ap: expalanations}.}\label{fig: htn_explanations}
\vspace{-0.5cm}
\end{figure}

In leveraging survival models for classification purposes, we have devised an alternative methodology to explain the decision-making process of these models. Our strategy is to develop a custom function that is designed to directly retrieve binary predictions from the survival model. Subsequently, we employ Shap's KernelExplainer on these predictions, thereby circumventing the need for intermediary surrogate models. Early comparisons of our method with SurvSHAP are revealed in Figure \ref{fig: htn_explanations}. They demonstrate the similarity between the explanations generated by both approaches. Notably, the concordance in the top five features, with four being identical across methods, and the alignment in feature importance for 18 of the top 20 features provide compelling evidence of the effectiveness of our approach. The consistency in the order of importance of features further validates the robustness and efficiency of our method.

The top features and risk factors identified for each disease were thoroughly vetted by our panel of clinical experts, who largely agreed with the observed trends.
\section{Limitations and Future Work}
Our study introduces a pioneering approach to survival analysis, striving for practical utility over traditional methods. One notable deviation from conventional survival analysis involves interpreting the survival curves depicted in Figure \ref{fig : surv_1}. Typically, the observed trend is not regarded as anomalous within the context of survival analysis. This trend could be attributed to the time distribution of samples of the diagnosed data. However, in a retrospective study, mitigating such biases presents significant challenges. Our methodology, designed to predict as early as possible for any patient, might diverge from traditional practices but the resulting models are clinically validated by our panel of physicians. By extending our analysis to normal examples beyond the censoring time, we enhanced the performance of survival models and obtained more intuitive survival curves.

 Further research directions include applying the random survival forest models to a broader spectrum of chronic diseases and investigating alternative strategies to enhance the performance of the models. Additionally, efforts will be made to adapt Shap's TreeExplainer for use with Random Survival Forests, which the library does not currently support.
\vspace{-0.5cm}
\section{Conclusion}
This study presents substantial advancements in healthcare predictive analytics, notably through the development of an early-warning, disease risk surveillance system for chronic diseases using survival models. The major contributions include creating a robust system that leverages routinely collected clinical EMR data excluding the labs to compute the disease risk of chronic diseases.

By integrating survival analysis with classification techniques, we've solved a notable challenge reported in healthcare research. This synthesis offers clinicians a unified model capable of providing multiple inferences, including disease prediction and risk assessment, thereby enabling timely and accurate clinical decision-making. Our approach not only streamlines the analytical process by using a singular model for risk predictions but also overcomes the limitations highlighted in past studies.

Our exploration into the realm of explainability further distinguishes our work, offering a novel methodology for explaining the decision-making processes of survival models without the need for intermediary surrogate models. The comparability of our explanations with those generated by established methods, such as SurvSHAP, validates the efficacy of our approach and marks a notable advancement in making complex models more transparent and interpretable for clinicians. Lastly, all features, models, disease risk predictions, and explanations are clinically validated by a panel of expert physicians. The clinical validation phase led to the selection of features considered clinically sound and hence the model training was guided by sound medical knowledge from the onset. Consequently, we tried our best to hold FAVES (Fair, Appropriate, Valid, Effective, and Safe) principle of predictive models for healthcare applications.

\bibliographystyle{splncs04}
\bibliography{bibliography}

@article{zou2018predicting,
  title={Predicting diabetes mellitus with machine learning techniques},
  author={Zou, Quan and Qu, Kaiyang and Luo, Yamei and Yin, Dehui and Ju, Ying and Tang, Hua},
  journal={Frontiers in genetics},
  volume={9},
  pages={515},
  year={2018},
  publisher={Frontiers Media SA}
}

@article{dinh2019data,
  title={A data-driven approach to predicting diabetes and cardiovascular disease with machine learning},
  author={Dinh, An and Miertschin, Stacey and Young, Amber and Mohanty, Somya D},
  journal={BMC medical informatics and decision making},
  volume={19},
  number={1},
  pages={1--15},
  year={2019},
  publisher={Springer}
}

@article{ilyas2021chronic,
  title={Chronic kidney disease diagnosis using decision tree algorithms},
  author={Ilyas, Hamida and Ali, Sajid and Ponum, Mahvish and Hasan, Osman and Mahmood, Muhammad Tahir and Iftikhar, Mehwish and Malik, Mubasher Hussain},
  journal={BMC nephrology},
  volume={22},
  number={1},
  pages={1--11},
  year={2021},
  publisher={BioMed Central}
}

@article{martinez2021review,
  title={A review of machine learning in hypertension detection and blood pressure estimation based on clinical and physiological data},
  author={Martinez-R{\'\i}os, Erick and Montesinos, Luis and Alfaro-Ponce, Mariel and Pecchia, Leandro},
  journal={Biomedical Signal Processing and Control},
  volume={68},
  pages={102813},
  year={2021},
  publisher={Elsevier}
}

@inproceedings{ahsan2022machine,
  title={Machine-learning-based disease diagnosis: A comprehensive review},
  author={Ahsan, Md Manjurul and Luna, Shahana Akter and Siddique, Zahed},
  booktitle={Healthcare},
  volume={10},
  number={3},
  pages={541},
  year={2022},
  organization={MDPI}
}

@article{petousis2024early,
  title={Early prediction of end-stage kidney disease using electronic health record data: a machine learning approach with a 2-year horizon},
  author={Petousis, Panayiotis and Wilson, James M and Gelvezon, Alex V and Alam, Shafiul and Jain, Ankur and Prichard, Laura and Elashoff, David A and Raja, Naveen and Bui, Alex AT},
  journal={JAMIA open},
  volume={7},
  number={1},
  pages={ooae015},
  year={2024},
  publisher={Oxford University Press}
}

@article{islam2023chronic,
  title={Chronic kidney disease prediction based on machine learning algorithms},
  author={Islam, Md Ariful and Majumder, Md Ziaul Hasan and Hussein, Md Alomgeer},
  journal={Journal of Pathology Informatics},
  volume={14},
  pages={100189},
  year={2023},
  publisher={Elsevier}
}

@inproceedings{mani2012type,
  title={Type 2 diabetes risk forecasting from EMR data using machine learning},
  author={Mani, Subramani and Chen, Yukun and Elasy, Tom and Clayton, Warren and Denny, Joshua},
  booktitle={AMIA annual symposium proceedings},
  volume={2012},
  pages={606},
  year={2012},
  organization={American Medical Informatics Association}
}

@inproceedings{liu2018early,
  title={Early prediction of diabetes complications from electronic health records: A multi-task survival analysis approach},
  author={Liu, Bin and Li, Ying and Sun, Zhaonan and Ghosh, Soumya and Ng, Kenney},
  booktitle={Proceedings of the AAAI Conference on Artificial Intelligence},
  volume={32},
  number={1},
  year={2018}
}

@article{hagar2014survival,
  title={Survival analysis with electronic health record data: Experiments with chronic kidney disease},
  author={Hagar, Yolanda and Albers, David and Pivovarov, Rimma and Chase, Herbert and Dukic, Vanja and Elhadad, No{\'e}mie},
  journal={Statistical Analysis and Data Mining: The ASA Data Science Journal},
  volume={7},
  number={5},
  pages={385--403},
  year={2014},
  publisher={Wiley Online Library}
}

@article{migora2021survival,
  title={Survival time to development of hypertension and its predictors among a cohort of diabetic patients in health facilities of gurage zone: A retrospective follow-up study},
  author={Migora, Biru and Geleso, Mulugeta Geremew and Girum, Tadele and Bireda, Meskele and Gebru, Mehari and Dessu, Samuel},
  journal={Vascular Health and Risk Management},
  pages={259--266},
  year={2021},
  publisher={Taylor \& Francis}
}

@article{matheson2018prediction,
  title={Prediction models for the development of COPD: a systematic review},
  author={Matheson, Melanie C and Bowatte, Gayan and Perret, Jennifer L and Lowe, Adrian J and Senaratna, Chamara V and Hall, Graham L and de Klerk, Nick and Keogh, Louise A and McDonald, Christine F and Waidyatillake, Nilakshi T and others},
  journal={International journal of chronic obstructive pulmonary disease},
  pages={1927--1935},
  year={2018},
  publisher={Taylor \& Francis}
}

@article{zhou2022development,
  title={Development of an electronic frailty index for predicting mortality and complications analysis in pulmonary hypertension using random survival forest model},
  author={Zhou, Jiandong and Chou, Oscar Hou In and Wong, Ka Hei Gabriel and Lee, Sharen and Leung, Keith Sai Kit and Liu, Tong and Cheung, Bernard Man Yung and Wong, Ian Chi Kei and Tse, Gary and Zhang, Qingpeng},
  journal={Frontiers in cardiovascular medicine},
  volume={9},
  pages={735906},
  year={2022},
  publisher={Frontiers}
}

@article{ahmad2023classification,
  title={Classification based on event in survival machine learning analysis of cardiovascular disease cohort},
  author={Ahmad, Shokh Mukhtar and Ahmed, Nawzad Muhammed},
  journal={BMC Cardiovascular Disorders},
  volume={23},
  number={1},
  pages={1--7},
  year={2023},
  publisher={BioMed Central}
}

@article{weiser2011predicting,
  title={Predicting survival after curative colectomy for cancer: individualizing colon cancer staging},
  author={Weiser, Martin R and G{\"o}nen, Mithat and Chou, Joanne F and Kattan, Michael W and Schrag, Deborah},
  journal={Journal of Clinical Oncology},
  volume={29},
  number={36},
  pages={4796},
  year={2011},
  publisher={American Society of Clinical Oncology}
}

@article{ye2018prediction,
  title={Prediction of incident hypertension within the next year: prospective study using statewide electronic health records and machine learning},
  author={Ye, Chengyin and Fu, Tianyun and Hao, Shiying and Zhang, Yan and Wang, Oliver and Jin, Bo and Xia, Minjie and Liu, Modi and Zhou, Xin and Wu, Qian and others},
  journal={Journal of medical Internet research},
  volume={20},
  number={1},
  pages={e22},
  year={2018},
  publisher={JMIR Publications Toronto, Canada}
}

@inproceedings{survSHAP,
  title={SurvSHAP: a proxy-based algorithm for explaining survival models with SHAP},
  author={Alabdallah, Abdallah and Pashami, Sepideh and R{\"o}gnvaldsson, Thorsteinn and Ohlsson, Mattias},
  booktitle={2022 IEEE 9th international conference on data science and advanced analytics (DSAA)},
  pages={1--10},
  year={2022},
  organization={IEEE}
}
\appendix
\newpage

\section{Feature Sets}\label{ap: features}

\subsection{Common Features}

\begin{table}[ht]
   \vspace{-1cm}
	\caption{Common Features included in all models}
	\label{tab : common_features}
    \centering
    \begin{tabular}{l|l}
 
    \hline
    Feature Name  & Feature Description \\
    \hline
    age\_group    &  Categorizes the age of a person into pre-defined interval or bins\\
    gender        &  Identifies the gender of a person\\
    race\_mapping &  A numerical representation of an individual's race\\
    bmi\_group    &  Categorizes the BMI of person into pre-defined interval or bins\\
    map\_group    &  Categorizes mean arterial pressure (MAP) of a person into pre-defined intervals or bins
    \vspace{-1cm}
    \end{tabular}
\end{table}

\subsection{Diabetes Specific Features}
\begin{table}[ht]
   \vspace{-1cm}
	\caption{Diabetes Features - Part 1}
	\label{tab : diabetes_features_1}
    \centering
    \begin{tabular}{l|l}
    \hline
    Feature Name  & Feature Description \\
    \hline
    E78      & Code for Disorders of lipoprotein metabolism and other lipidemias (Present/Absent)     \\
    I25      & Code for Chronic ischemic heart disease (Present/Absent)     \\
    K21      & Code for Gastro-esophageal reflux disease (Present/Absent)     \\
    K29      & Code for Gastritis and duodenitis (Present/Absent)     \\
    N17      & Code for Acute kidney failure (Present/Absent)     \\
    R60      & Code for Edema, not elsewhere classified (Present/Absent)     \\ 
    F41      & Code for Other anxiety disorders (Present/Absent)     \\
    E10      & Code for Type 1 diabetes mellitus (Present/Absent)     \\
    N18      & Code for Chronic kidney disease (CKD) (Present/Absent)     \\
    CHF      & Elixhauser Comorbidity Group for Congestive heart failure (Present/Absent) \\
    HTN      & Elixhauser Comorbidity Group for Primary hypertension (Present/Absent)     \\
    D12.6     & Code for Benign neoplasm of colon, unspecified (Present/Absent)     \\
    ARTH     & Elixhauser Comorbidity Group for Rheumatoid arthritis (Present/Absent)     \\
    DRUG     & Elixhauser Comorbidity Group for Drug abuse (Present/Absent)     \\
    VALVE    & Elixhauser Comorbidity Group for Valvular disease (Present/Absent)     \\
    ULCER    & Elixhauser Comorbidity Group for Chronic peptic ulcer disease (Present/Absent)     \\
    TUMOR   & Elixhauser Comorbidity Group for Solid tumor without metastasis (Present/Absent)      \\
    LYTES   & Elixhauser Comorbidity Group for Fluid and electrolyte disorders (Present/Absent)      \\
    HYPOTHY & Elixhauser Comorbidity Group for Hypothyroidism (Present/Absent)      \\
    ANEMDEF & Elixhauser Comorbidity Group for Deficiency anemias (Present/Absent)      \\
    statins & Medications under GPI code 3940 (Prescribed/Not Prescribed)      \\
    PERIVASC& Elixhauser Comorbidity Group for Peripheral vascular disease (Present/Absent)      \\
    progestin& Medications under GPI code 2600 (Prescribed/Not Prescribed)      \\
    dibenzapines& Medications under GPI code 5915 (Prescribed/Not Prescribed)   \\   
    \end{tabular}
       \vspace{-1cm}
\end{table}

\begin{table}[ht]
	\caption{Diabetes Features - Part 2}
	\label{tab : diabetes_features_2}
    \centering
    \begin{tabular}{l|l}
    \hline
    Feature Name  & Feature Description \\
    \hline
    dexamethasone& Medications under GPI code 2210 (Prescribed/Not Prescribed)  \\
    sleep\_disorder& Custom Feature Group for codes  G47, F51 and Z72.820 (Present/Absent)\\
    horomonal\_antineoplastic& Medications under GPI code 2140 (Prescribed/Not Prescribed)    \\
    smoking\_history & Custom Feature Group indicating smoking history (Present/Absent)  \\
    family\_diabetes & Family History of Diabetes (Present/Absent)  \\
    metabolic\_disorder& Custom Feature Group for codes E88 and Z86.3 (Present/Absent) \\
    gestational\_diabetes    & Custom Feature Group for codes O24.4 and Z86.32 (Present/Absent)  \\
    \end{tabular}
\end{table}

\subsection{Hypertension Specific Features}
\begin{table}[ht]
\vspace{-1cm}
	\caption{Hypertension Features}
	\label{tab : hypertension_features}
    \centering
    \begin{tabular}{p{0.3\textwidth}|l}
    \hline
    Feature Name  & Feature Description \\
    \hline
    E11      & Code for Type 2 diabetes mellitus (Present/Absent)    \\
    E78      & Code for Disorders of lipoprotein metabolism and other lipidemias (Present/Absent)     \\
    G89      & Code for Pain, not elsewhere classified (Present/Absent)     \\ 
    I25      & Code for Chronic ischemic heart disease (Present/Absent)     \\
    F41      & Code for Other anxiety disorders (Present/Absent)     \\
    K29      & Code for Gastritis and duodenitis (Present/Absent)    \\
    R06      & Code for Abnormalities of breathing (Present/Absent)     \\
    R07      & Code for Pain in throat and chest (Present/Absent)     \\
    J20      & Code for Acute bronchitis (Present/Absent)     \\
    K30      & Code for Functional dyspepsia (Present/Absent)     \\ 
    L40      & Code for Psoriasis (Present/Absent)     \\
    gerd     & Custom Feature Group for codes K21 and Z87.19 (Present/Absent)     \\
    nsaids      & Medications under GPI code 6610 (Prescribed/Not Prescribed)      \\
    E88.810      & Code for Metabolic syndrome (Present/Absent)    \\
    family\_heart     & Family History of Heart diseases (Present/Absent)    \\ 
    marital\_status     & Indicator of Marital status (Married/Unmarried)     \\
    osteoarthritis     & Custom Feature Group for codes  (Present/Absent)     \\ 
    sleep\_disorder     & Custom Feature Group for codes  G47, F51 and Z72.820 (Present/Absent)     \\
    duration\_of\_E66     & Duration in years from earliest occurrence of code for Overweight and obesity    \\ 
    duration\_of\_R73     &  Duration in years from earliest occurrence of code for Elevated blood glucose level \\
    family\_hypertension      &  Family History of Hypertensive diseases (Present/Absent)    \\
    reductase\_inhibitors     & Medications under GPI code 3940 (Prescribed/Not Prescribed)      \\
    rheumatoid\_arthritis      & Custom Feature Group for codes M05 and M06  (Present/Absent)     \\
    chronic\_lung\_disease      & Custom Feature Group for codes J41-J45 (Present/Absent)     \\
    
    thyroid\_gland\_disorders      & Custom Feature Group for codes E02-E06 (Present/Absent)     \\
    serotonin\_norepinephrine reuptake\_inhibitors  &  Medications under GPI code 5818 (Prescribed/Not Prescribed)     \\
    \end{tabular}
    \vspace{-1cm}
\end{table}
\clearpage
\subsection{CKD Specific Features}
\begin{table}[ht]
	\caption{CKD Features}
	\label{tab : kidney_features}
    \centering
    \begin{tabular}{l|l}
    \hline
    Feature Name  & Feature Description \\
    \hline
    E11      & Code for Type 2 diabetes mellitus (Present/Absent)     \\
    D50      & Code for Iron deficiency anemia (Present/Absent)     \\
    R53      & Code for Malaise and fatigue (Present/Absent)     \\ 
    I25      & Code for Chronic ischemic heart disease (Present/Absent)     \\
    R06      & Code for Abnormalities of breathing (Present/Absent)     \\
    R73      & Code for Elevated blood glucose level (Present/Absent)     \\
    G47      & Code for Sleep disorders (Present/Absent)     \\
    E05      & Code for Thyrotoxicosis [hyperthyroidism] (Present/Absent)     \\
    HTN      & Elixhauser Comorbidity Group for Primary hypertension (Present/Absent)     \\
    E87.6      & Code for Hypokalemia (Present/Absent)     \\ 
    gerd     &  Custom Feature Group for codes K21 and Z97.19 (Present/Absent)    \\
    HTNCX      & Elixhauser Comorbidity Group for Hypertensive encephalopathy (Present/Absent)     \\
    TUMOR      & Elixhauser Comorbidity Group for Solid tumor without metastasis (Present/Absent)     \\
    nsaids      &  Medications under GPI code 3940 (Prescribed/Not Prescribed)     \\
    CHRNLUNG      & Elixhauser Comorbidity Group for Chronic pulmonary disease (Present/Absent)     \\
    PERIVASC     & Elixhauser Comorbidity Group for Peripheral vascular disease (Present/Absent)     \\ 
    antagonists     &  Medications under GPI code 3615 (Prescribed/Not Prescribed)     \\
    family\_kidney     & Family History of Kidney diseases (Present/Absent) \\ 
    osteoarthritis     &  Custom Feature Group for codes M15-M19 (Present/Absent)    \\
    loop\_diuretics     &  Medications under GPI code 3720 (Prescribed/Not Prescribed)     \\ 
    family\_hypertension      & Family History of Hypertensive diseases (Present/Absent)     \\
    hypercholesterolemia     & Custom Feature Group for codes E78.0, E78.2, E78.4 and E78.5 (Present/Absent)     \\
    proton\_pump\_inhibitors      & Medications under GPI code 4927 (Prescribed/Not Prescribed)      \\
    \end{tabular}
\end{table}
\clearpage
\subsection{Heart Specific Features}
\begin{table}[ht]
\vspace{-1cm}
	\caption{Heart Features}
	\label{tab : heart_features}
    \centering
    \begin{tabular}{l|l}
    \hline
    Feature Name  & Feature Description \\
    \hline
    DM      &  Elixhauser Comorbidity Group for Diabetes without chronic complications (Present/Absent)    \\
    D50      & Code for Iron deficiency anemia (Present/Absent)    \\
    HTN      & Elixhauser Comorbidity Group for Primary hypertension (Present/Absent)    \\ 
    E78      & Code for Disorders of lipoprotein metabolism and other lipidemias (Present/Absent)    \\
    R51      & Code for Headache (Present/Absent)     \\
    R06      & Code for Abnormalities of breathing (Present/Absent)     \\
    R53      & Code for Malaise and fatigue (Present/Absent)     \\
    R07      & Code for Pain in throat and chest (Present/Absent)     \\
    I48      & Code for Atrial fibrillation and flutter (Present/Absent)     \\
    R42      & Code for Dizziness and giddiness (Present/Absent)     \\ 
    DMCX      & Elixhauser Comorbidity Group for Diabetes with chronic complications (Present/Absent)     \\
    gerd     &  Custom Feature Group for codes K21 and Z97.19 (Present/Absent)    \\
    HTNCX      & Elixhauser Comorbidity Group for Hypertensive encephalopathy (Present/Absent)     \\
    nsaids      & Medications under GPI code 6610 (Prescribed/Not Prescribed)      \\
    PERIVASC     & Elixhauser Comorbidity Group for Peripheral vascular disease (Present/Absent)     \\ 
    RENLFAIL     & Elixhauser Comorbidity Group for Renal failure (Present/Absent)     \\
    depression      & Custom Feature Group for codes/groups F32 and DEPRESS (Present/Absent)     \\
    hepatitis\_c      &  Custom Feature Group for codes B17.1, B18.2 and B19.2 (Present/Absent)    \\
    family\_heart      & Family History of Heart diseases (Present/Absent)     \\
    varicose\_veins      & Custom Feature Group for codes I83, O22.0, and O87.4 (Present/Absent)     \\ 
    osteoarthritis      &  Custom Feature Group for codes M15-M19 (Present/Absent)    \\
    smoking\_history     & Custom feature group indicating smoking history (Present/Absent)     \\
    rheumatoid\_arthritis      & Custom Feature Group for codes M05.0-M05.2, M05.4-M05.9, and M06 (Present/Absent)     \\
    chronic\_lung\_disease      & Custom Feature Group for codes/groups J41-J45 and CHRNLUNG (Present/Absent)     \\
    proton\_pump\_inhibitors     & Medications under GPI code 4927 (Prescribed/Not Prescribed)      \\ 
    aortic\_valve\_disorders     & Custom Feature Group for codes/groups I06, I35, and VALVE (Present/Absent)     \\
    neurological\_disorders      & Custom Feature Group for codes G11, G47, G43, and F51 (Present/Absent)     \\
    thyroid\_gland\_disorders     & Custom Feature Group for codes E00-E07 and E89.0 (Present/Absent)     \\ 
    \end{tabular}
    \vspace{-1cm}
\end{table}
\clearpage
\subsection{COPD Specific Features}
\begin{table}[ht]
\vspace{-1cm}
	\caption{COPD Features}
	\label{tab : copd_features}
    \centering
    \begin{tabular}{l|l}
    \hline
    Feature Name  & Feature Description \\
    \hline
    DM      & Elixhauser Comorbidity Group for Diabetes without chronic complications (Present/Absent)     \\
    R07      & Code for Pain in throat and chest (Present/Absent)     \\
    R09      & Code for Other symptoms and signs involving the circulatory and respiratory system (Present/Absent)     \\ 
    R53      & Code for Malaise and fatigue (Present/Absent)     \\
    E78      & Code for Disorders of lipoprotein metabolism and other lipidemias (Present/Absent)     \\
    F41      & Code for Other anxiety disorders (Present/Absent)     \\
    F03      & Code for Unspecified dementia (Present/Absent)     \\
    I25      & Code for Chronic ischemic heart disease (Present/Absent)     \\
    J45      & Code for Asthma (Present/Absent)     \\ 
    J30      & Code for Vasomotor and allergic rhinitis (Present/Absent)    \\
    J06      &  Code for Acute upper respiratory infections of multiple and unspecified sites (Present/Absent)    \\
    J02      & Code for Acute pharyngitis (Present/Absent)     \\
    J01      & Code for Acute sinusitis (Present/Absent)     \\
    HTN     & Elixhauser Comorbidity Group for Primary hypertension (Present/Absent)     \\ 
    CHF     & Elixhauser Comorbidity Group for Congestive heart failure (Present/Absent)     \\
    gerd      &  Custom Feature Group for codes K21 and Z97.19 (Present/Absent)    \\ 
    DMCX      & Elixhauser Comorbidity Group for Diabetes with chronic complications (Present/Absent)     \\
    DRUG      & Elixhauser Comorbidity Group for Drug abuse (Present/Absent)     \\
    J84.10      & Code for Pulmonary fibrosis, unspecified (Present/Absent)     \\
    TUMOR      & Elixhauser Comorbidity Group for Solid tumor without metastasis (Present/Absent)     \\
    LYTES      & Elixhauser Comorbidity Group for Fluid and electrolyte disorders (Present/Absent)     \\ 
    PSYCH      & Elixhauser Comorbidity Group for Psychoses (Present/Absent)     \\
    DEPRESS      & Elixhauser Comorbidity Group for Depression (Present/Absent)     \\
    ANEMDEF      & Elixhauser Comorbidity Group for Deficiency anemias (Present/Absent)    \\
    HYPOTHY     & Elixhauser Comorbidity Group for Hypothyroidism (Present/Absent)     \\ 
    wheezing     &  Custom Feature Group for codes  (Present/Absent)    \\
    PERIVASC      & Elixhauser Comorbidity Group for Peripheral vascular disease (Present/Absent)     \\
    RENLFAIL      & Elixhauser Comorbidity Group for  (Present/Absent)     \\
    respiration      & Identifies respiration rate of a person \\
    family\_lung      & Family History of Lung diseases (Present/Absent)     \\
    cough\_and\_cold      &  Custom Feature Group for codes R05, and A37 (Present/Absent)    \\
    sleep\_disorder     & Custom Feature Group for codes G47, F51 and Z72.820 (Present/Absent)     \\
    smoking\_history      & Custom feature group indicating smoking history (Present/Absent)     \\
    \end{tabular}
    \vspace{-2cm}
\end{table}

\clearpage

\section{Time Distribution Plots}\label{ap: time_distributions}

\begin{figure}
\vspace{-0.5cm}
	\centering
 \begin{subfigure}[b]{\textwidth}
	\begin{subfigure}[b]{0.32\textwidth}
		\includegraphics[width=\textwidth]{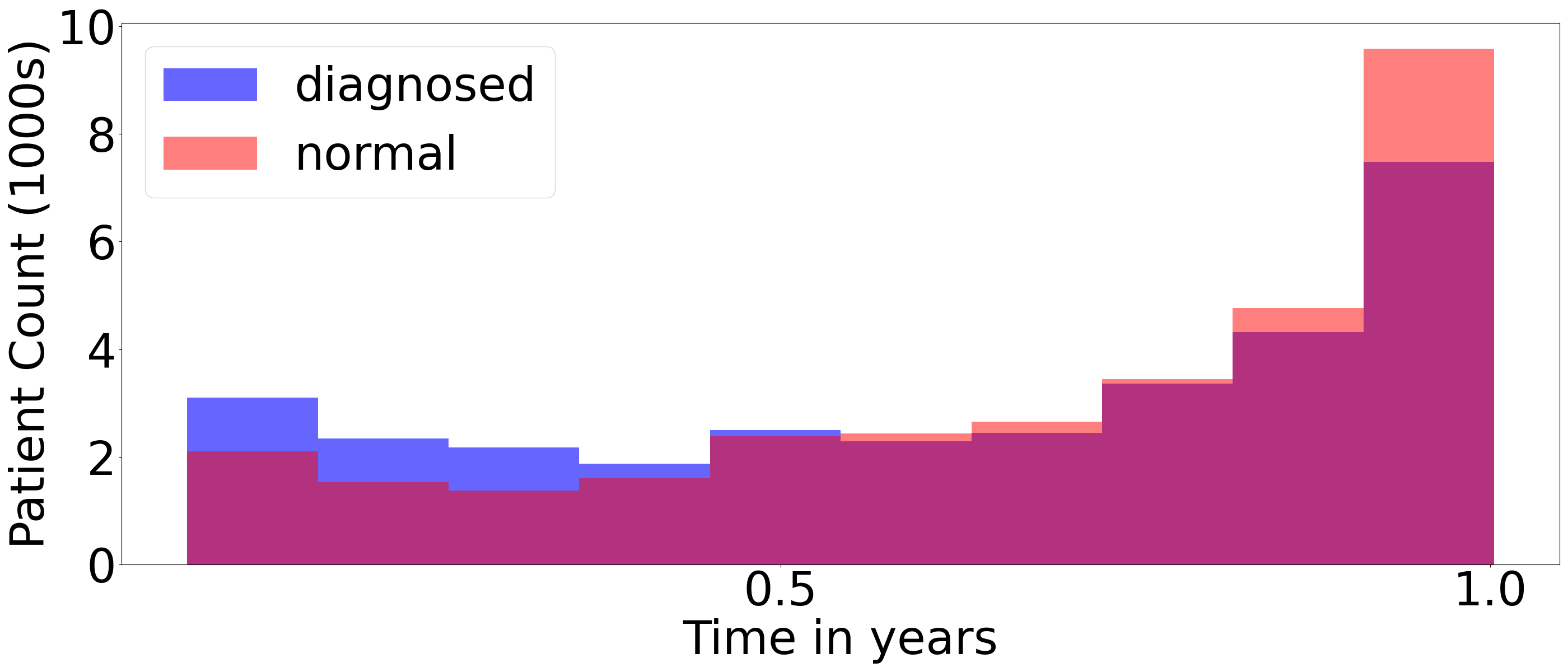}
		\caption{Approach 1: Similar}
	\end{subfigure}
\begin{subfigure}[b]{0.32\textwidth}
	\includegraphics[width=\textwidth]{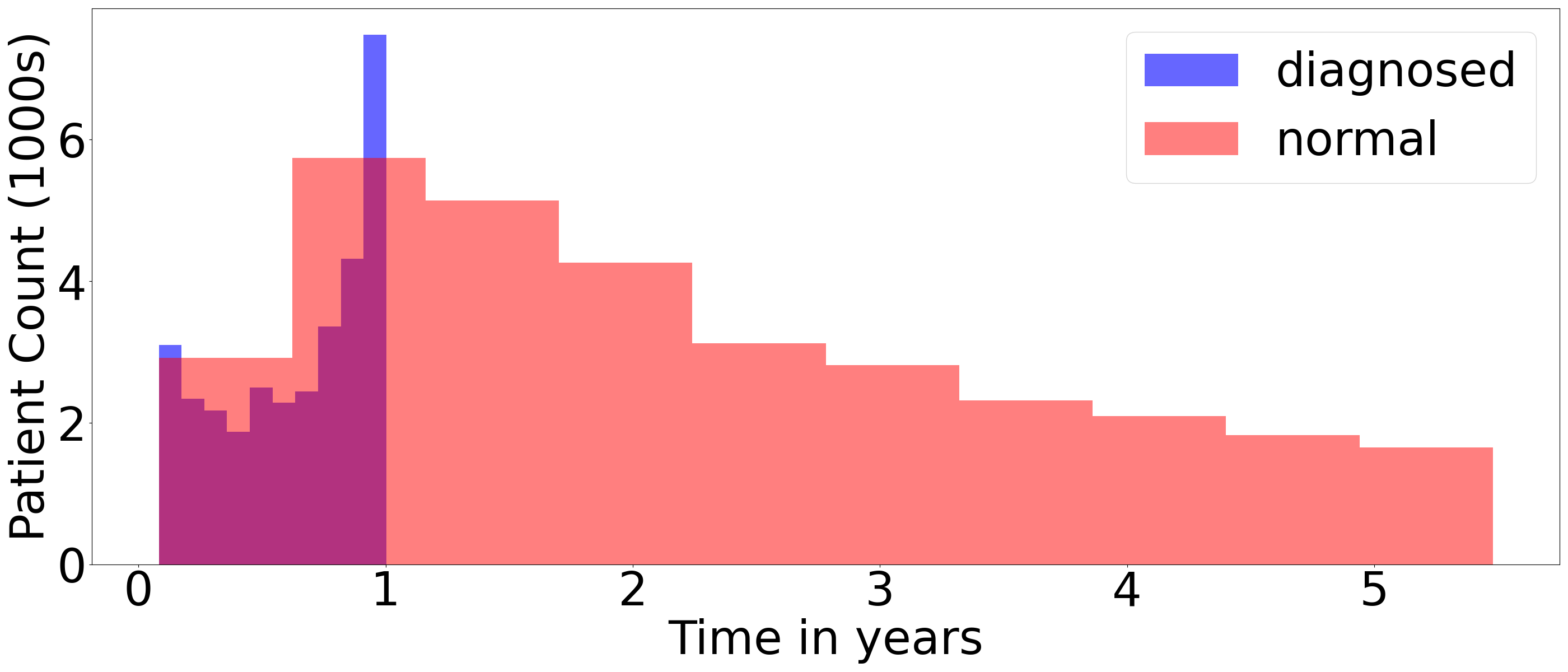}
	\caption{Approach 2: Overlap} 
\end{subfigure}
\begin{subfigure}[b]{0.32\textwidth}
	\includegraphics[width=\textwidth]{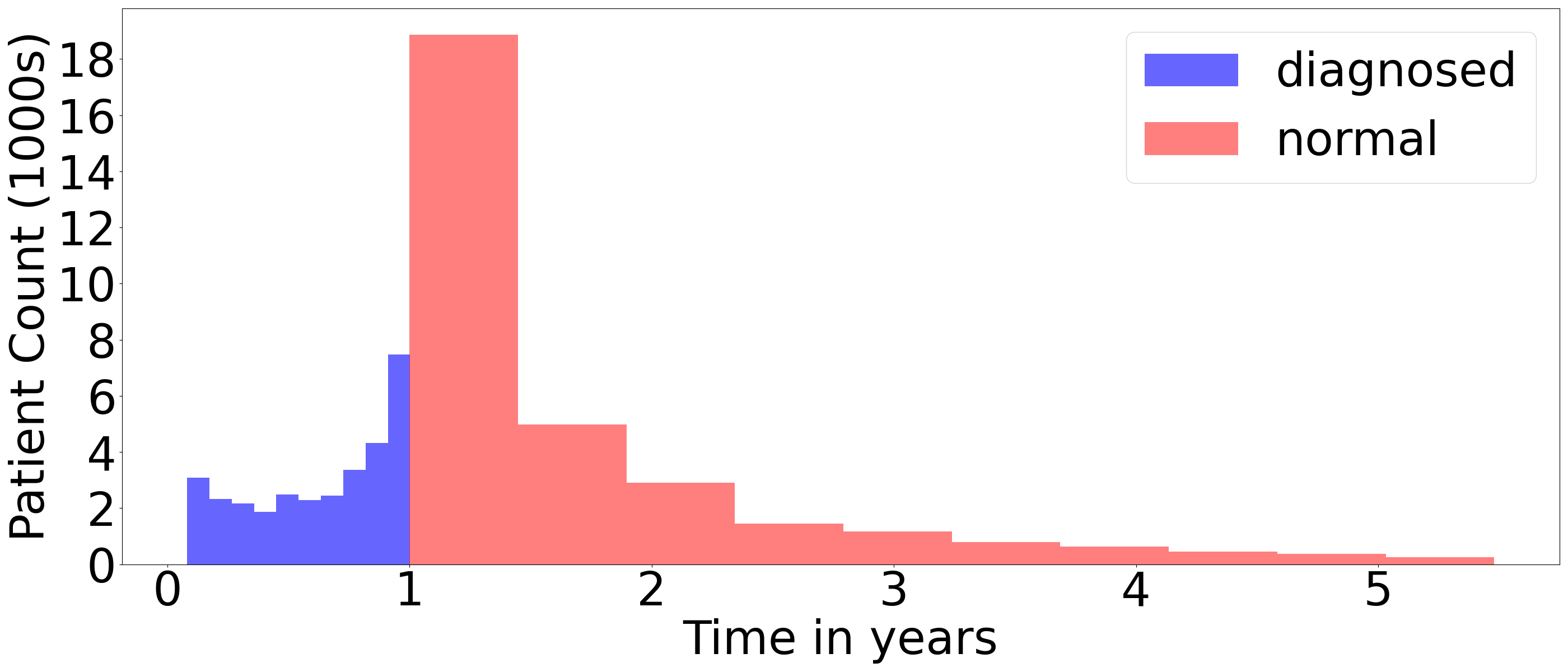}
	\caption{Approach 3: Distinct}
\end{subfigure}
	\caption{Diabetes}

\end{subfigure}
 \begin{subfigure}[b]{\textwidth}
	\begin{subfigure}[b]{0.32\textwidth}
		\includegraphics[width=\textwidth]{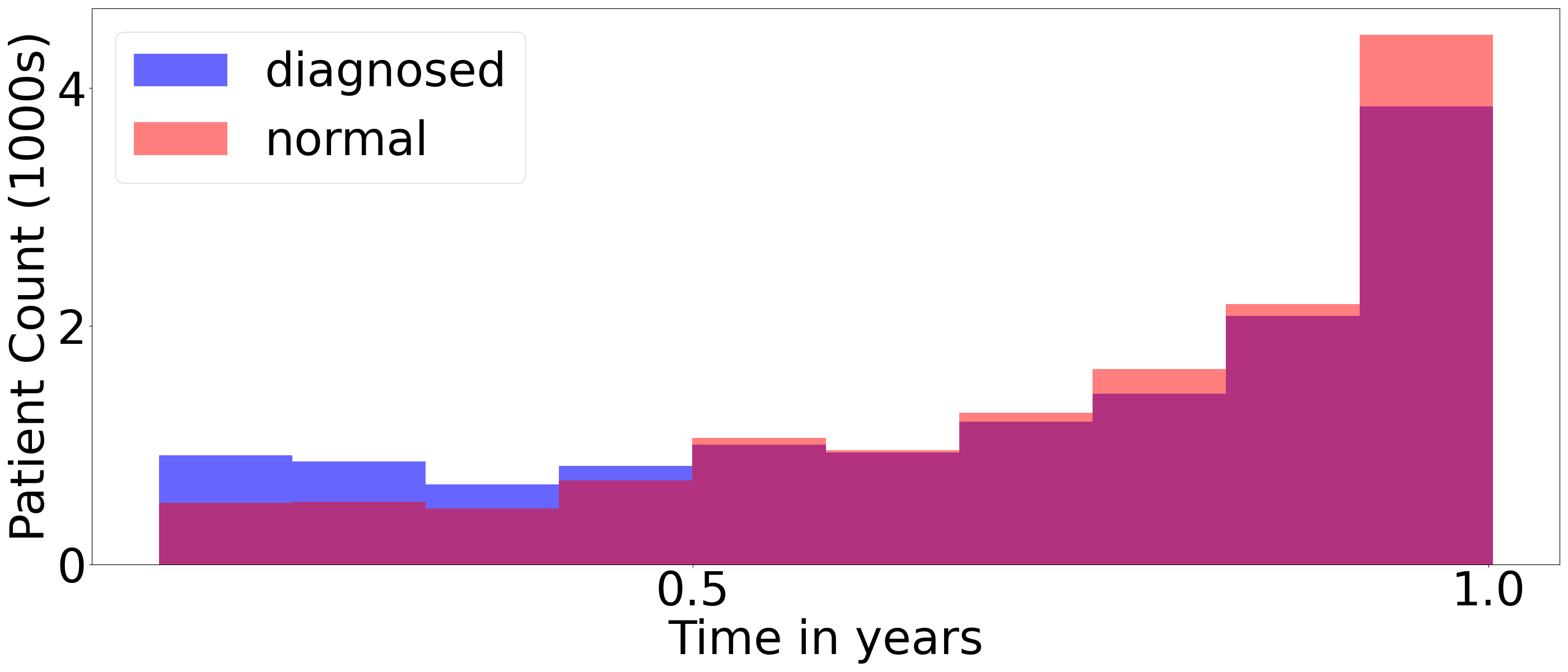}
		\caption{Approach 1: Similar}
	\end{subfigure}
\begin{subfigure}[b]{0.32\textwidth}
	\includegraphics[width=\textwidth]{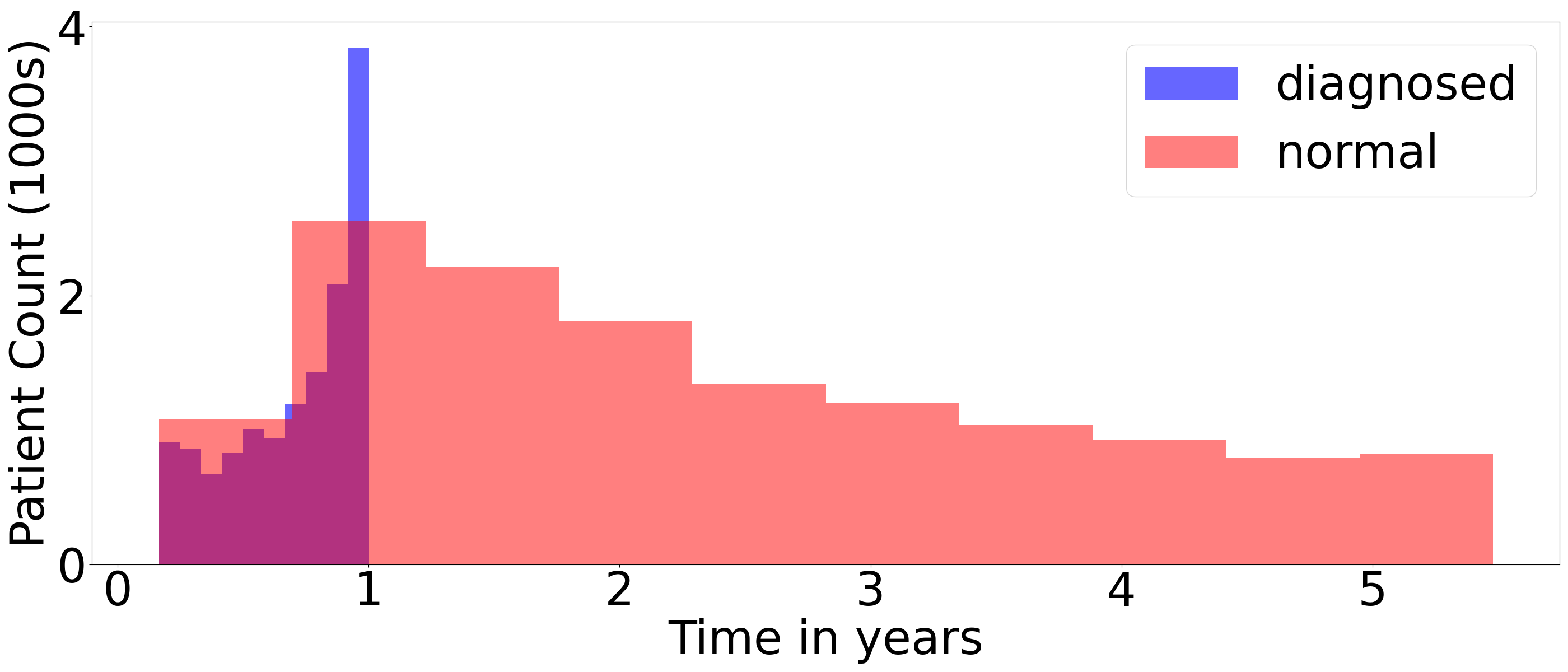}
	\caption{Approach 2: Overlap} 
\end{subfigure}
\begin{subfigure}[b]{0.32\textwidth}
	\includegraphics[width=\textwidth]{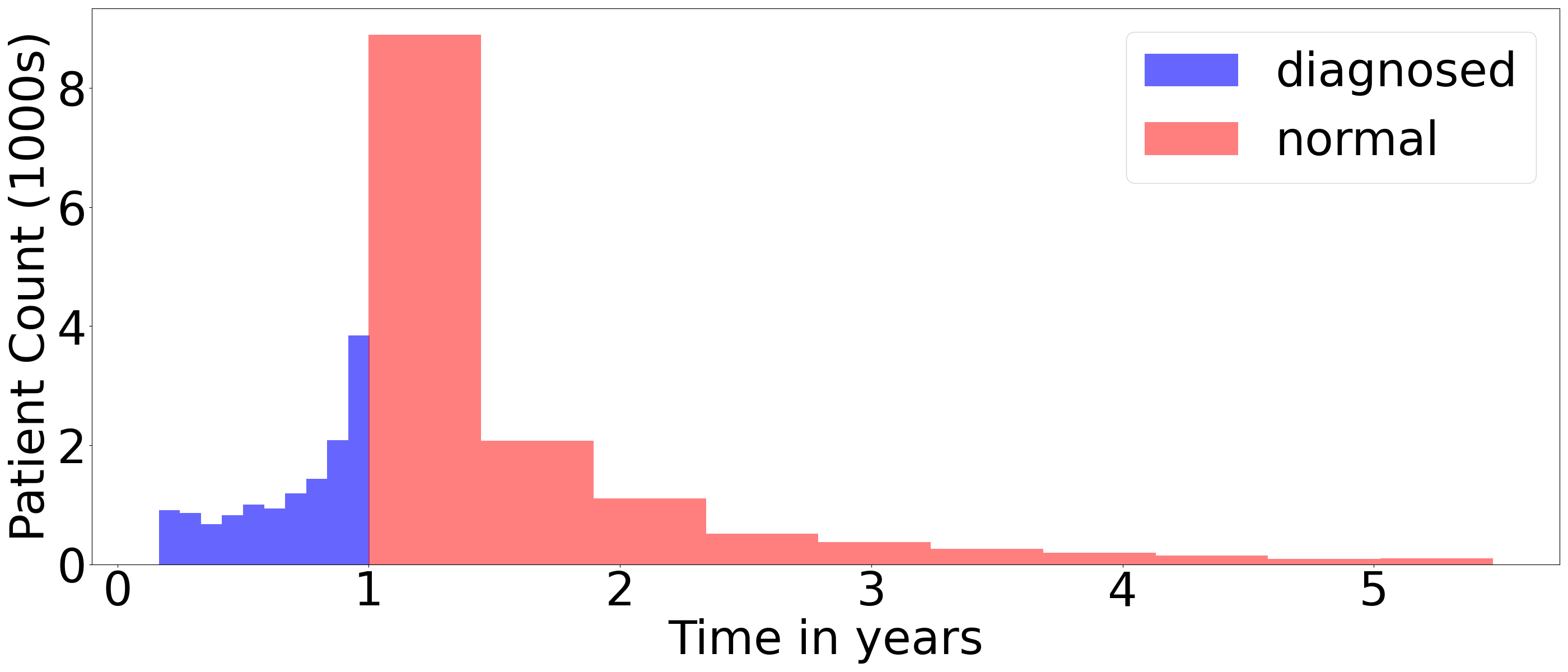}
	\caption{Approach 3: Distinct}
\end{subfigure}
	\caption{Heart}

\end{subfigure}
 \begin{subfigure}[b]{\textwidth}
	\begin{subfigure}[b]{0.32\textwidth}
		\includegraphics[width=\textwidth]{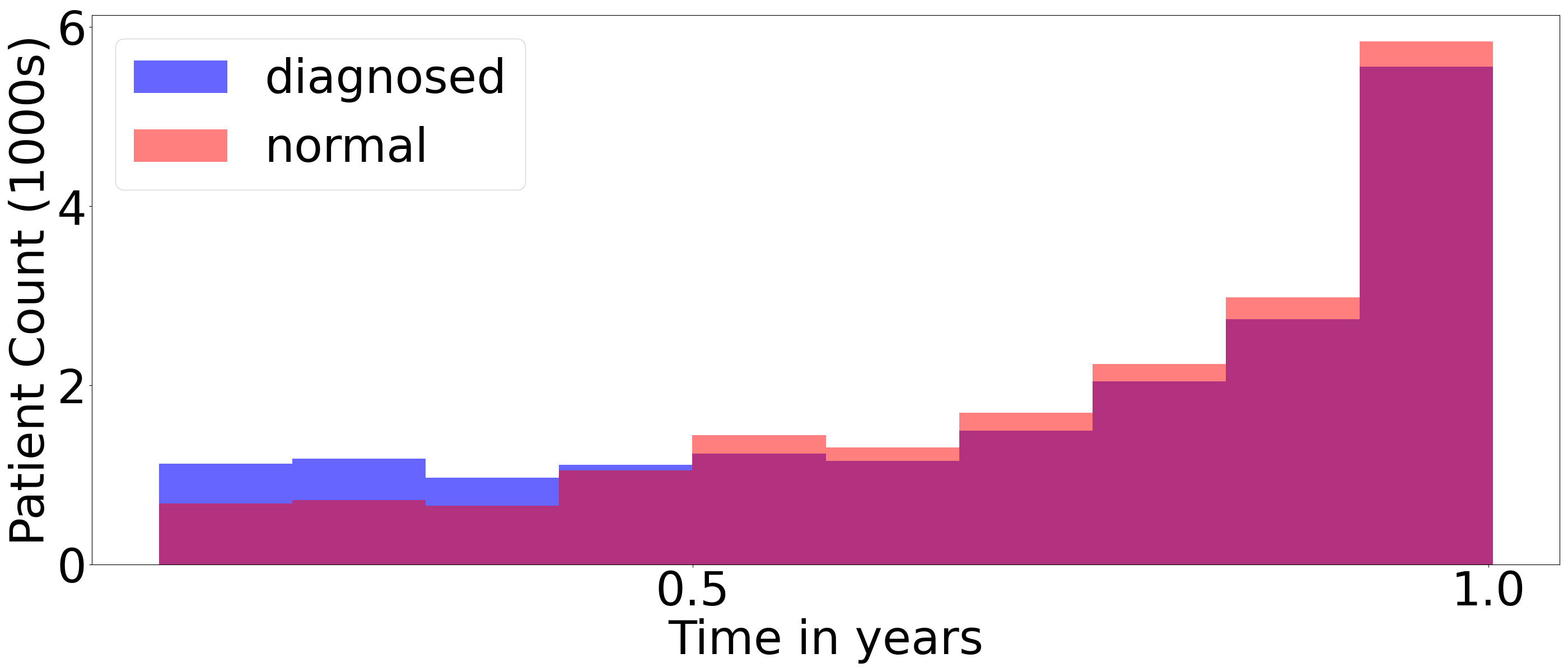}
		\caption{Approach 1: Similar}
	\end{subfigure}
\begin{subfigure}[b]{0.32\textwidth}
	\includegraphics[width=\textwidth]{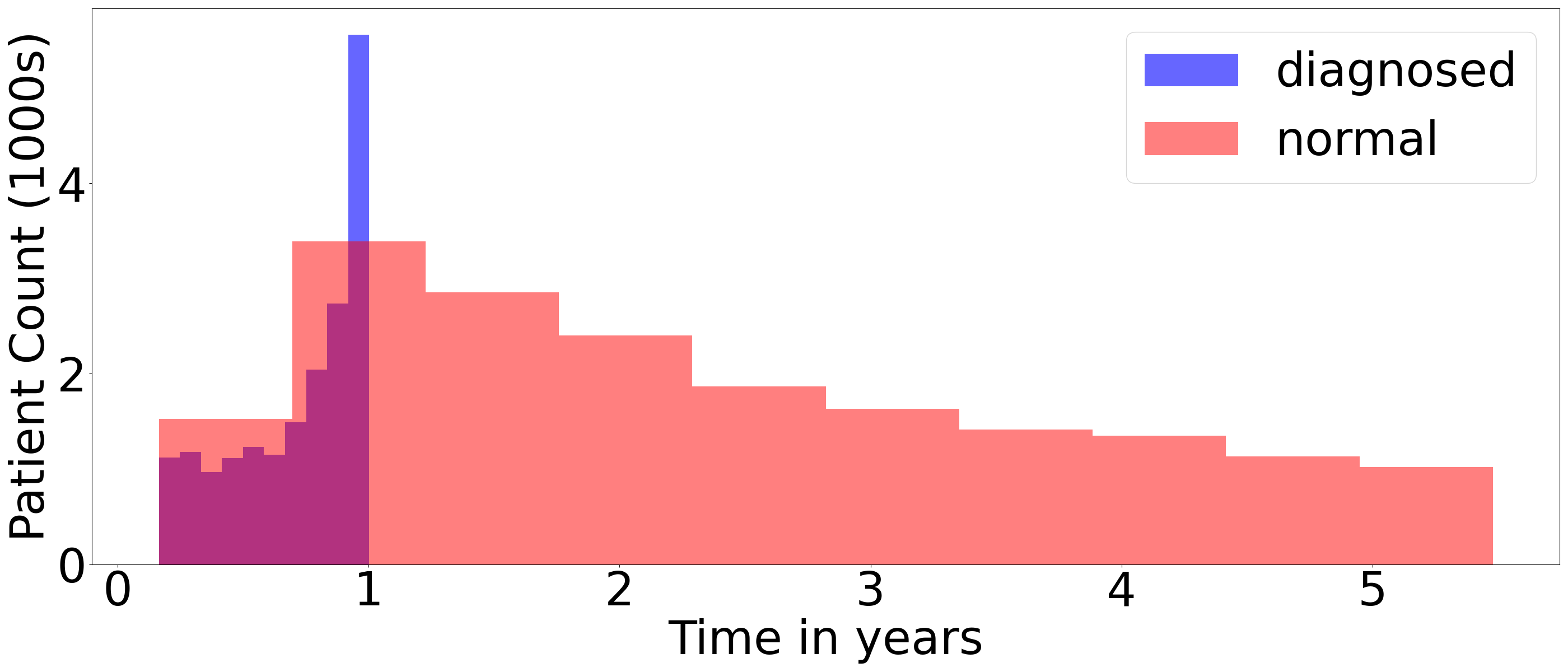}
	\caption{Approach 2: Overlap} 
\end{subfigure}
\begin{subfigure}[b]{0.32\textwidth}
	\includegraphics[width=\textwidth]{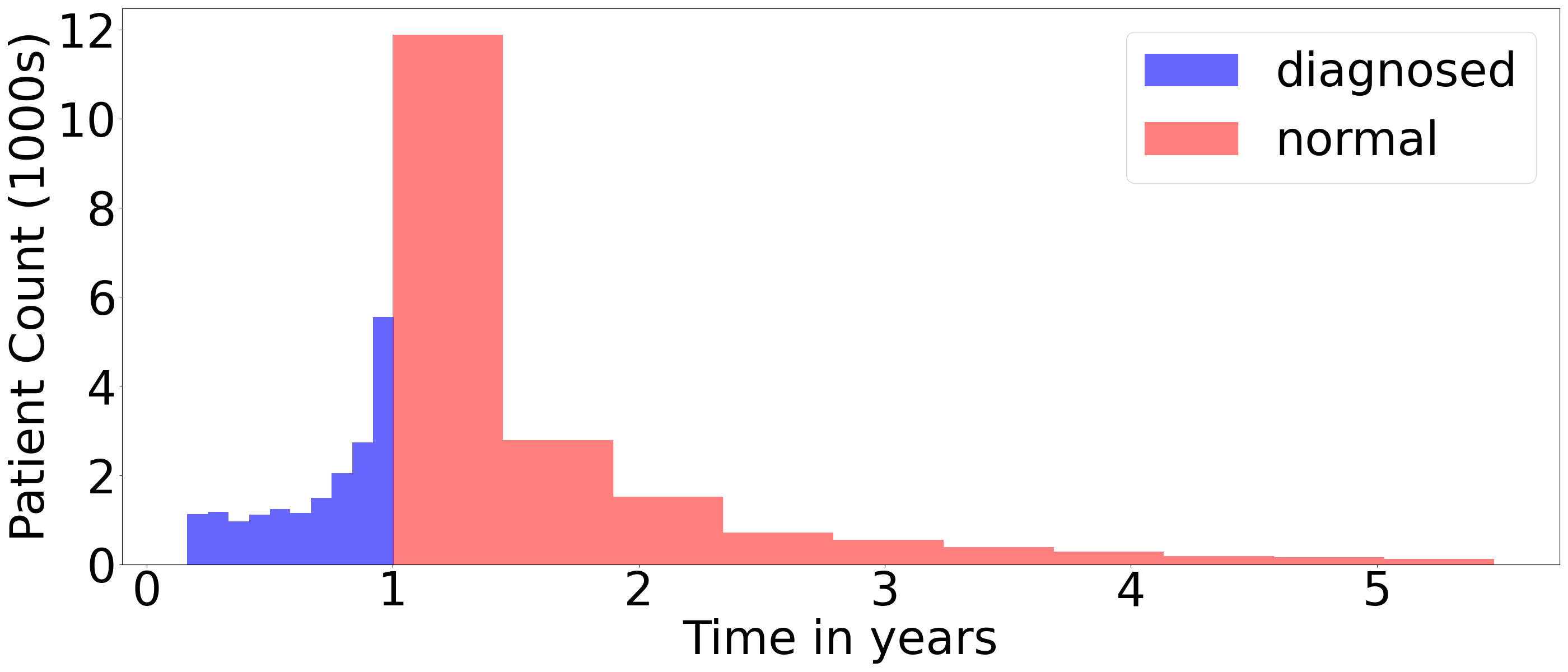}
	\caption{Approach 3: Distinct}
\end{subfigure}
	\caption{CKD}

\end{subfigure}
 \begin{subfigure}[b]{\textwidth}
	\begin{subfigure}[b]{0.32\textwidth}
		\includegraphics[width=\textwidth]{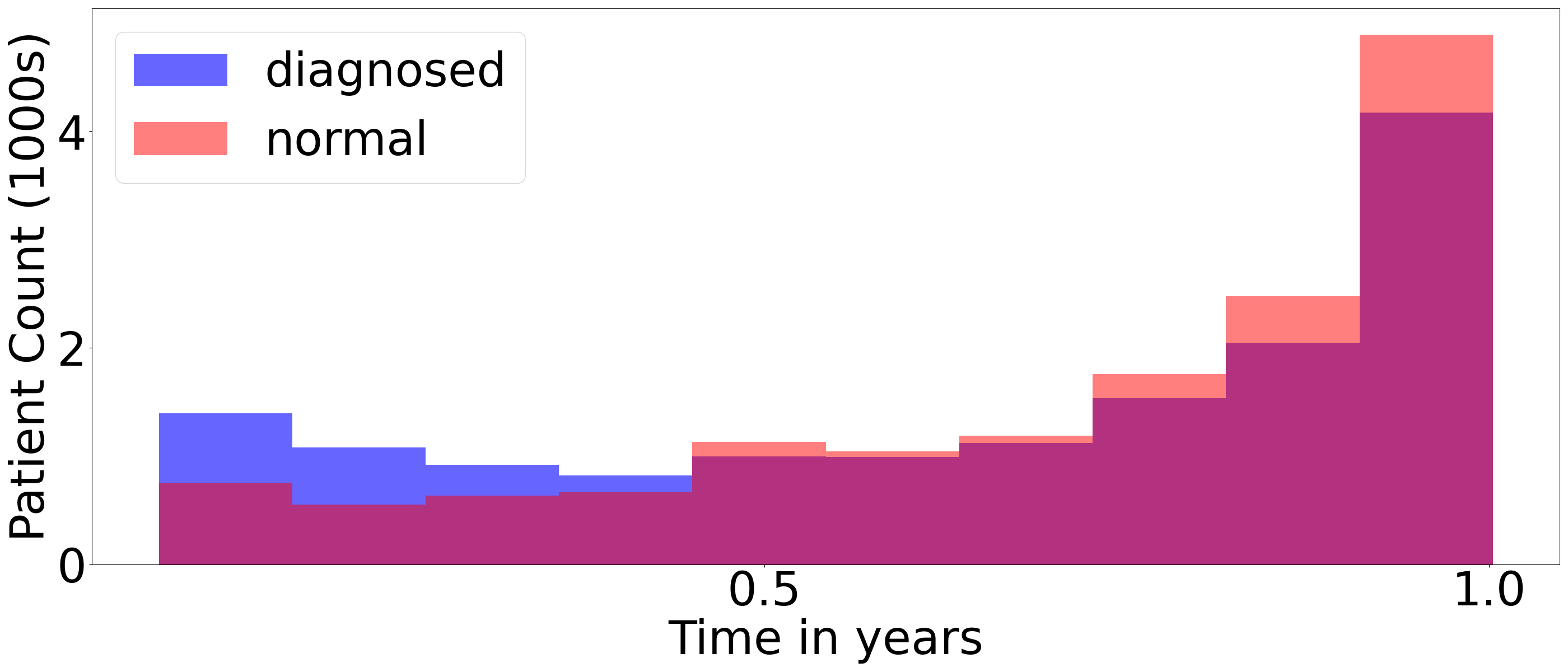}
		\caption{Approach 1: Similar}
	\end{subfigure}
\begin{subfigure}[b]{0.32\textwidth}
	\includegraphics[width=\textwidth]{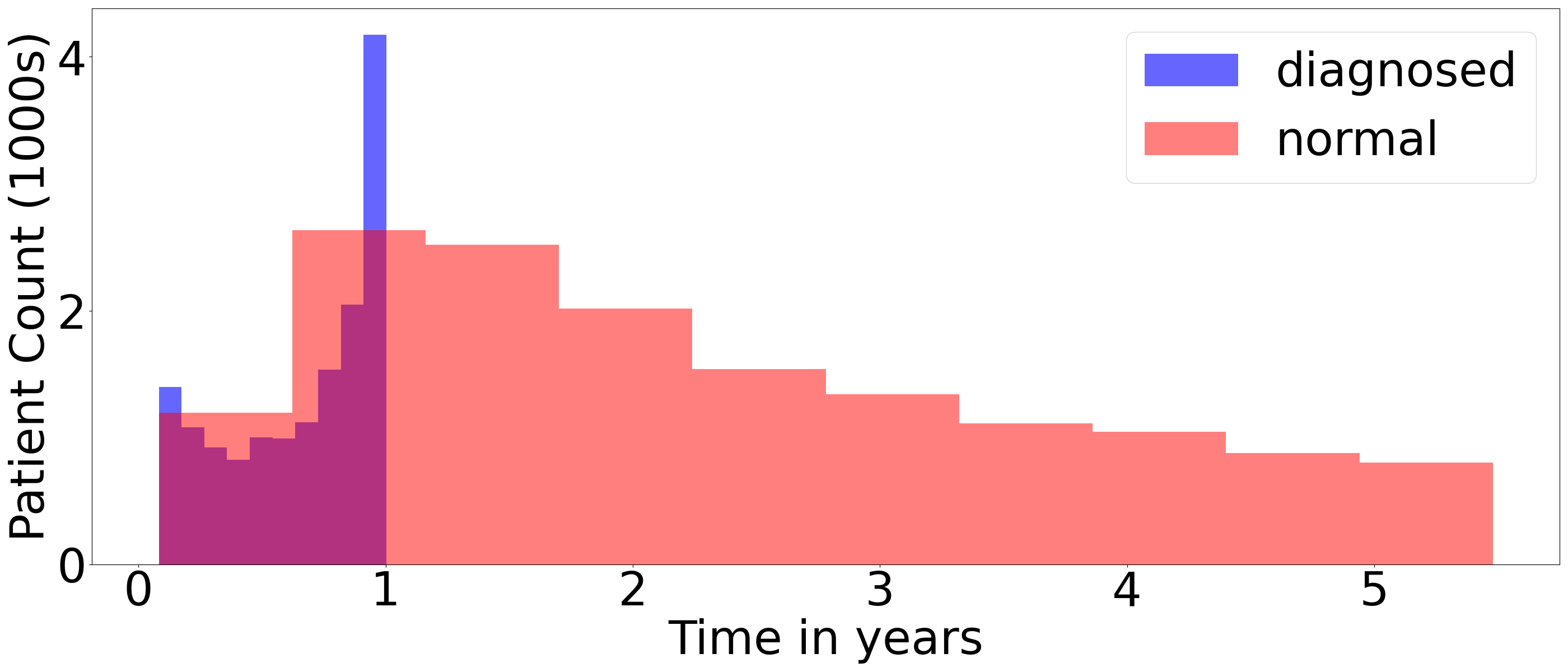}
	\caption{Approach 2: Overlap} 
\end{subfigure}
\begin{subfigure}[b]{0.32\textwidth}
	\includegraphics[width=\textwidth]{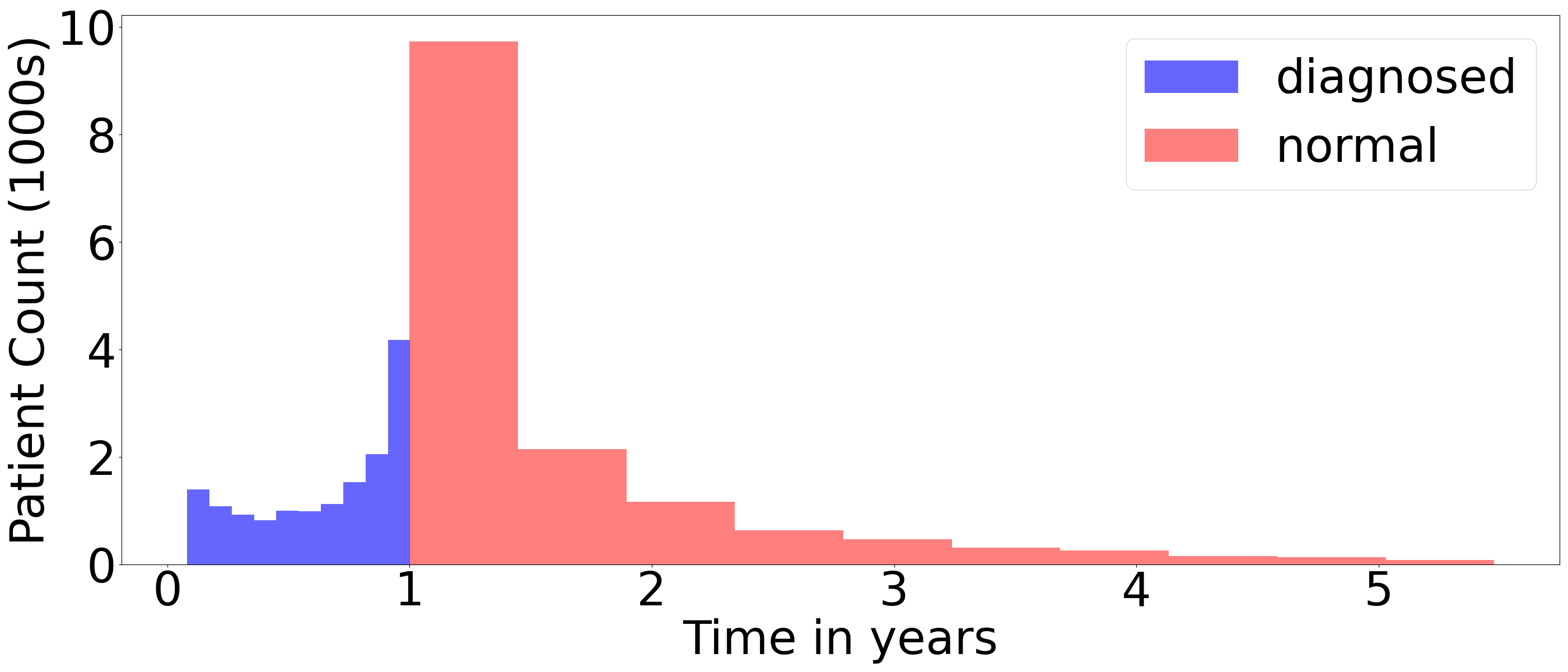}
	\caption{Approach 3: Distinct}
\end{subfigure}
	\caption{COPD}

\end{subfigure}
\end{figure}

\section{Clustered Survival Curve Analysis}\label{ap: survival_curves}

\begin{figure}
\vspace{-0.5cm}
	\centering
 \begin{subfigure}[b]{\textwidth}
	\begin{subfigure}[b]{0.32\textwidth}
		\includegraphics[width=\textwidth]{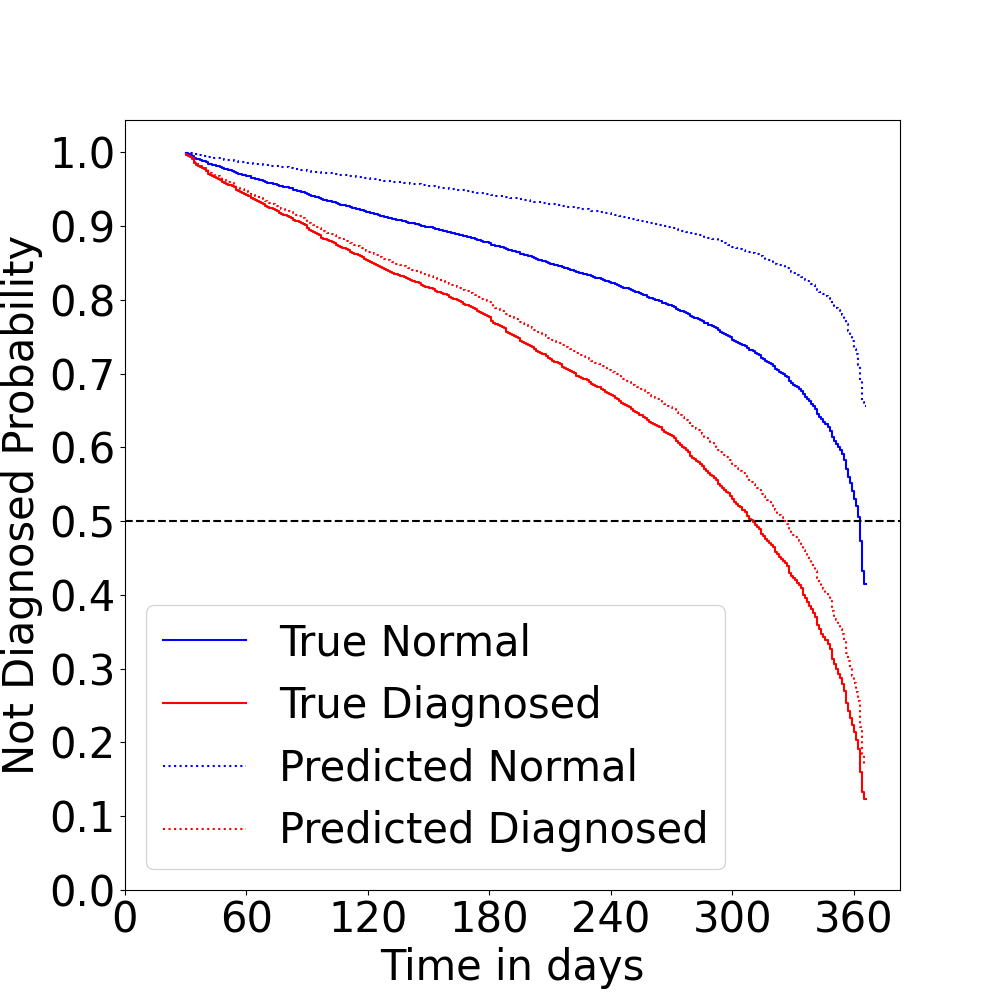}
		\caption{Approach 1: Similar}
	\end{subfigure}
\begin{subfigure}[b]{0.32\textwidth}
	\includegraphics[width=\textwidth]{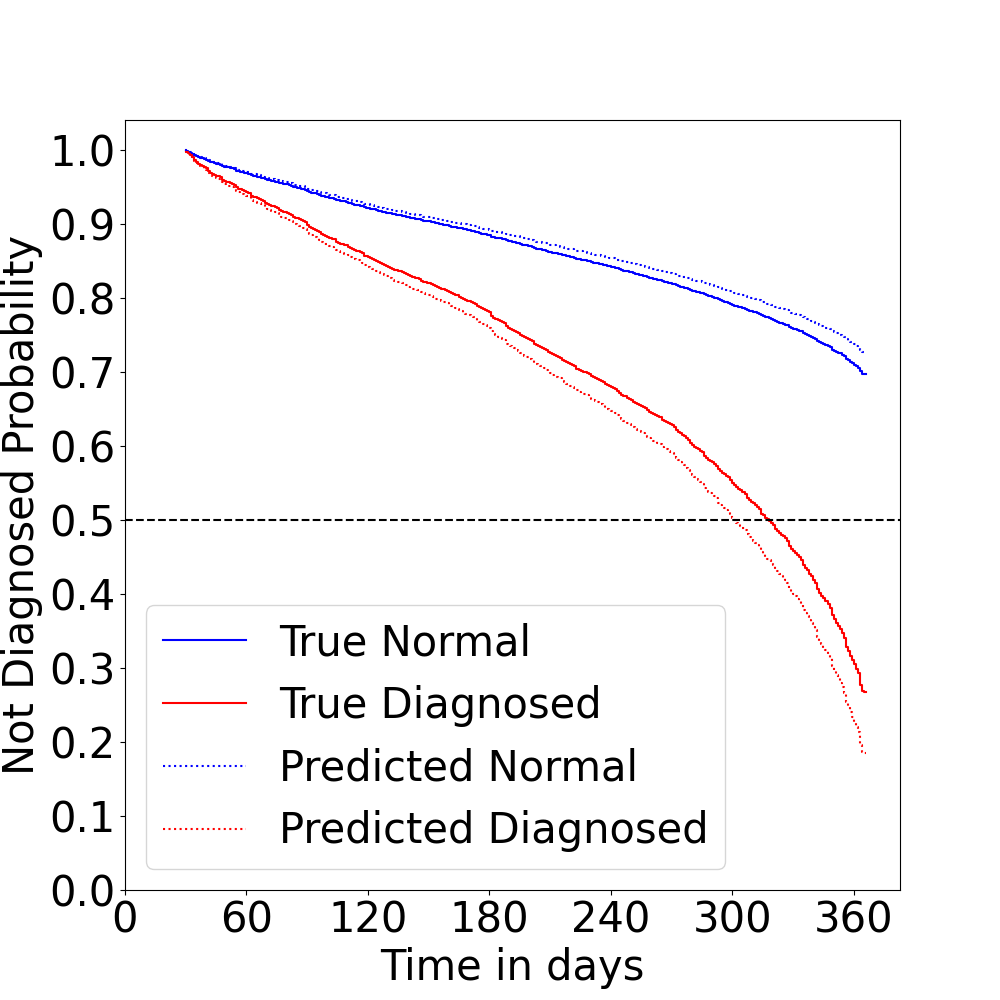}
	\caption{Approach 2: Overlap} 
\end{subfigure}
\begin{subfigure}[b]{0.32\textwidth}
	\includegraphics[width=\textwidth]{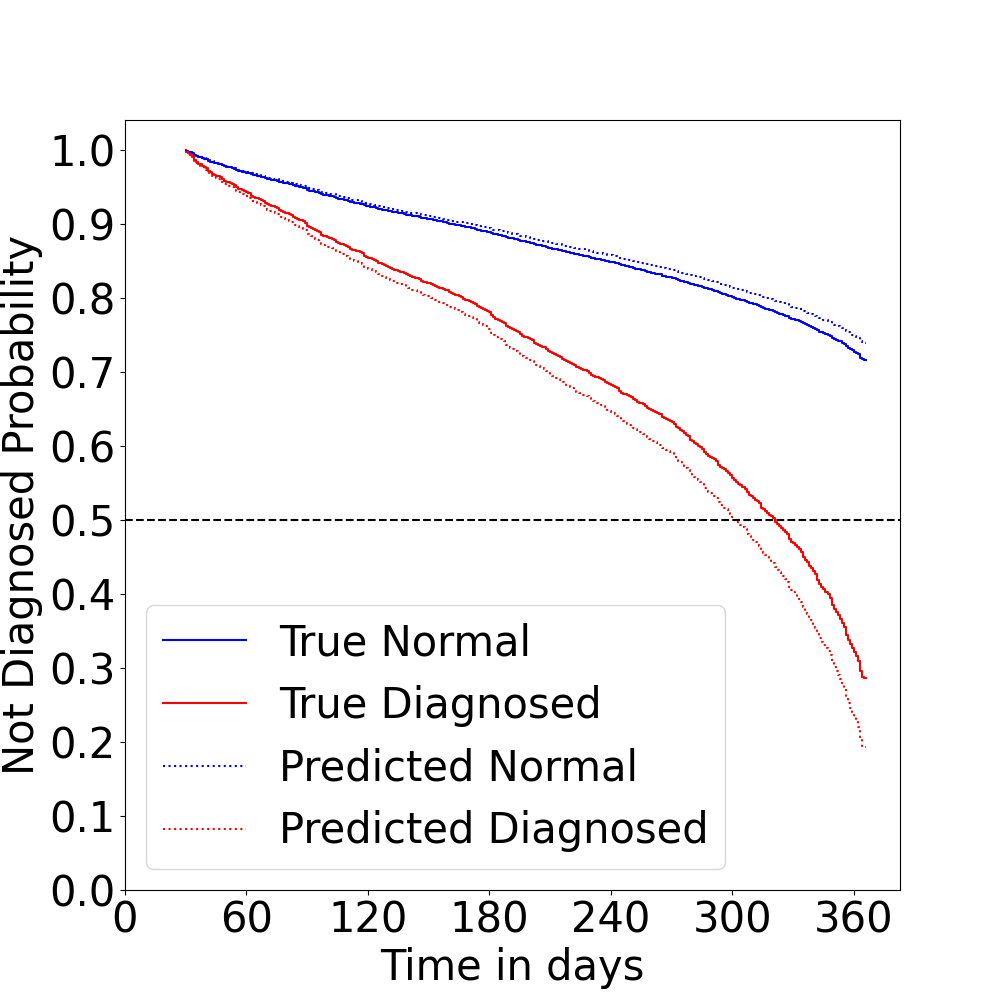}
	\caption{Approach 3: Distinct}
\end{subfigure}
	\caption{Diabetes}

\end{subfigure}
 \begin{subfigure}[b]{\textwidth}
	\begin{subfigure}[b]{0.32\textwidth}
		\includegraphics[width=\textwidth]{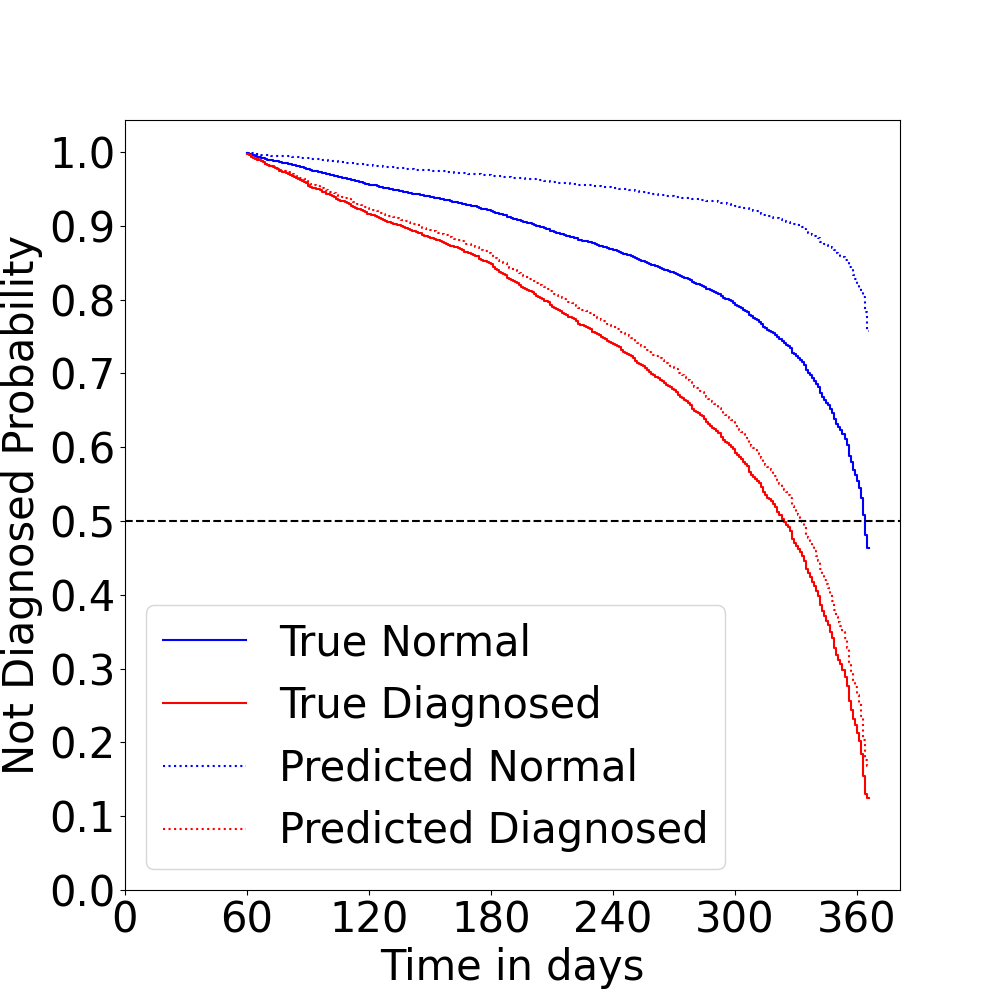}
		\caption{Approach 1: Similar}
	\end{subfigure}
\begin{subfigure}[b]{0.32\textwidth}
	\includegraphics[width=\textwidth]{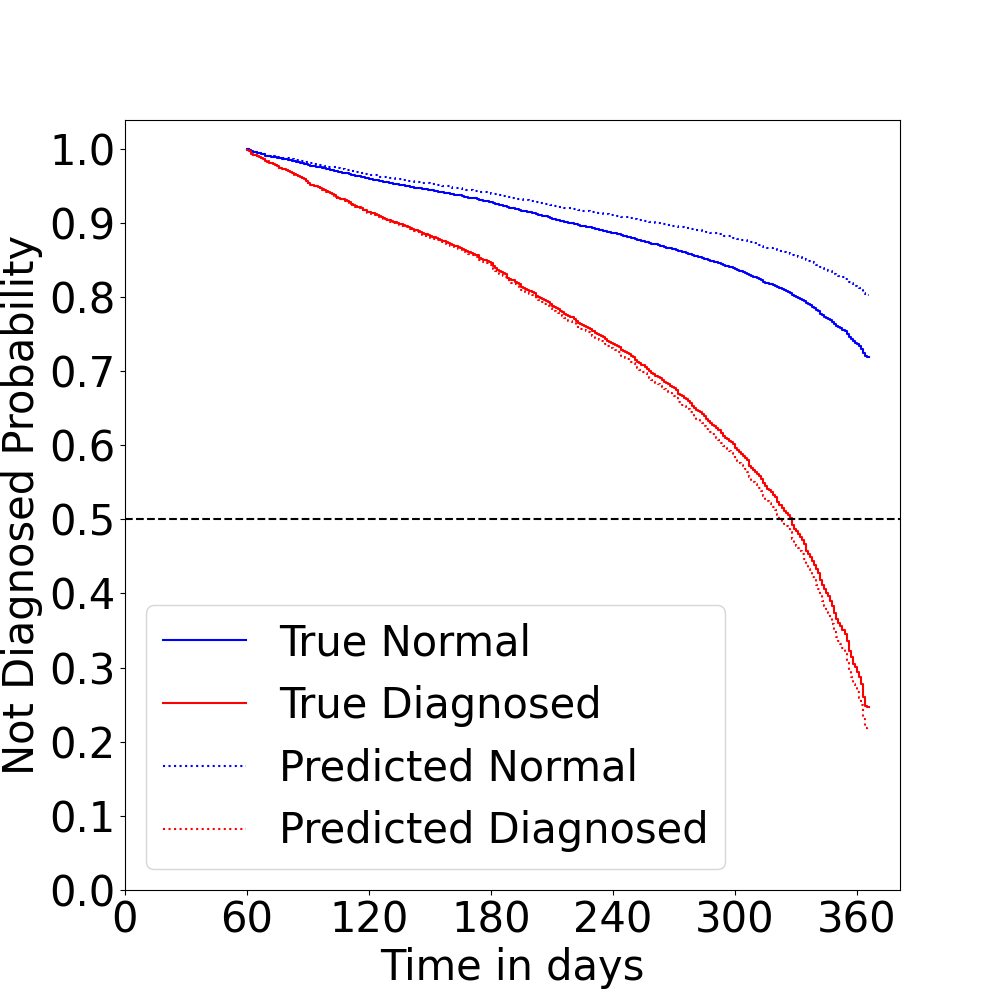}
	\caption{Approach 2: Overlap} 
\end{subfigure}
\begin{subfigure}[b]{0.32\textwidth}
	\includegraphics[width=\textwidth]{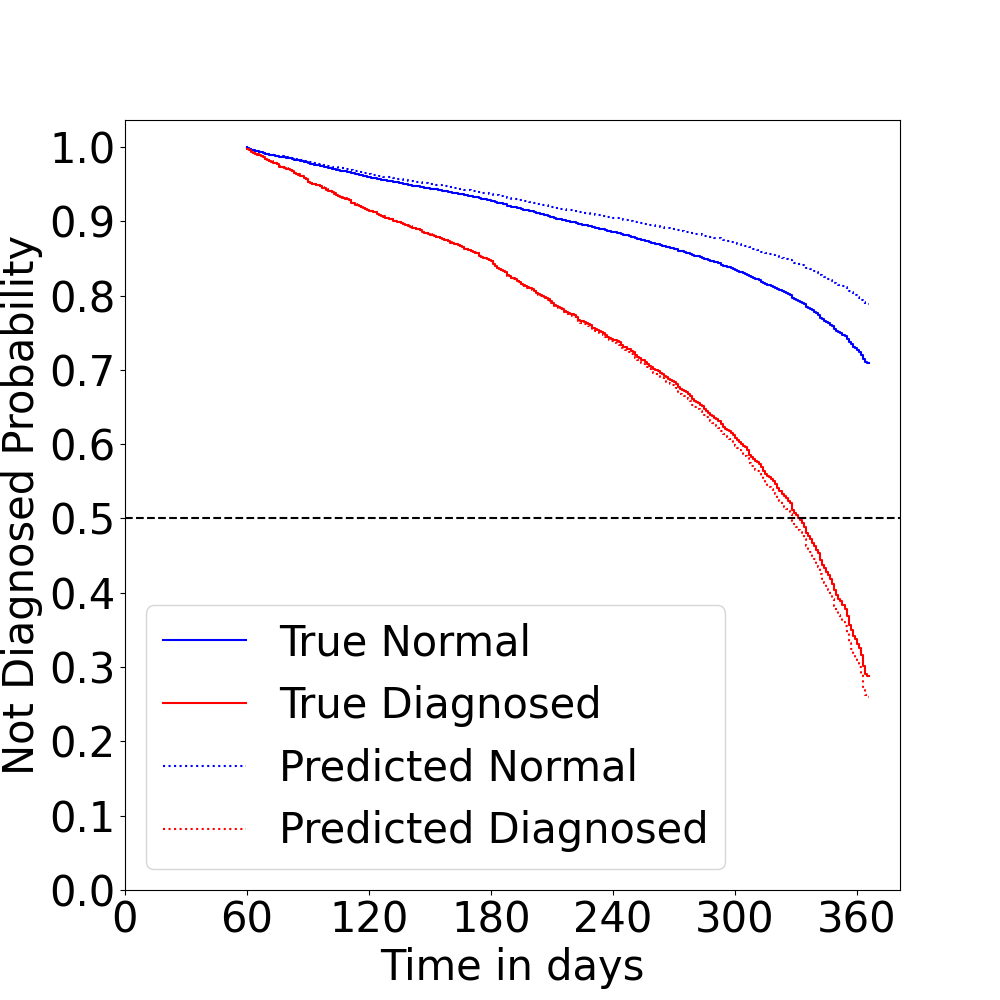}
	\caption{Approach 3: Distinct}
\end{subfigure}
	\caption{Heart}

\end{subfigure}
 \begin{subfigure}[b]{\textwidth}
	\begin{subfigure}[b]{0.32\textwidth}
		\includegraphics[width=\textwidth]{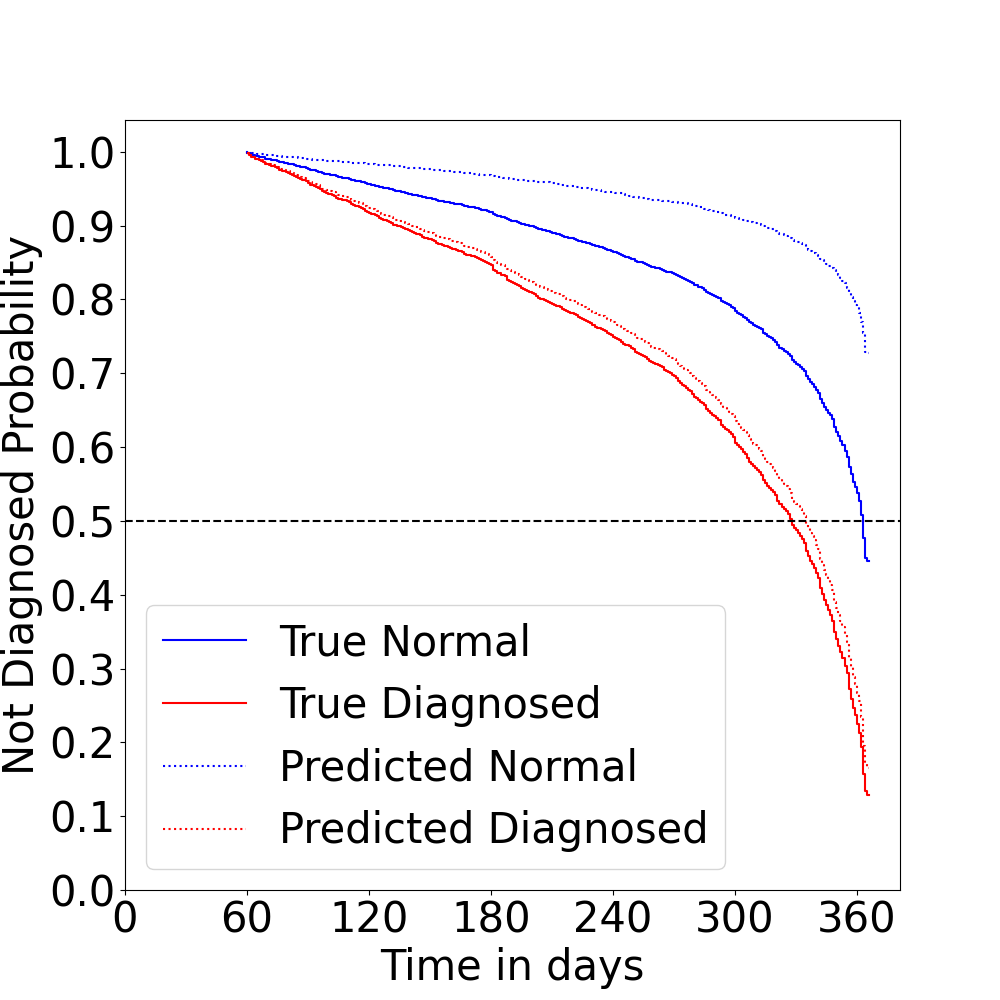}
		\caption{Approach 1: Similar}
	\end{subfigure}
\begin{subfigure}[b]{0.32\textwidth}
	\includegraphics[width=\textwidth]{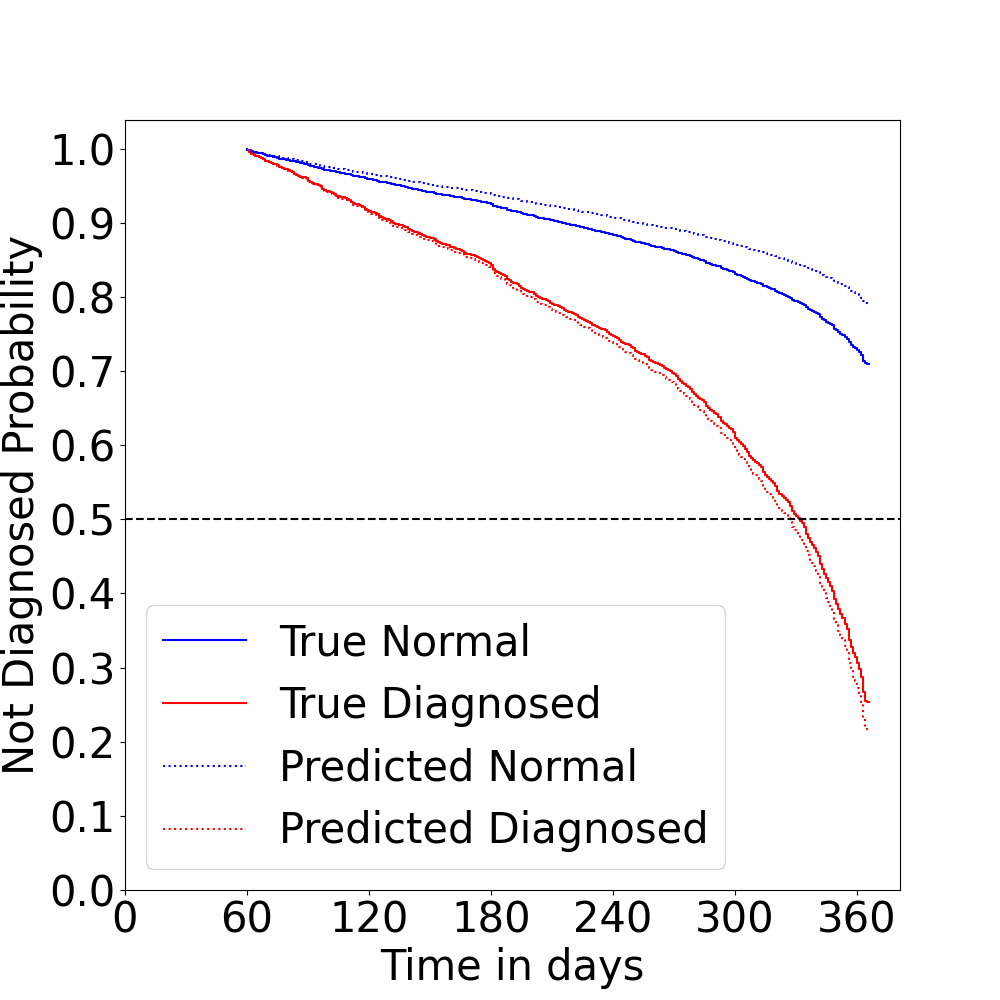}
	\caption{Approach 2: Overlap} 
\end{subfigure}
\begin{subfigure}[b]{0.32\textwidth}
	\includegraphics[width=\textwidth]{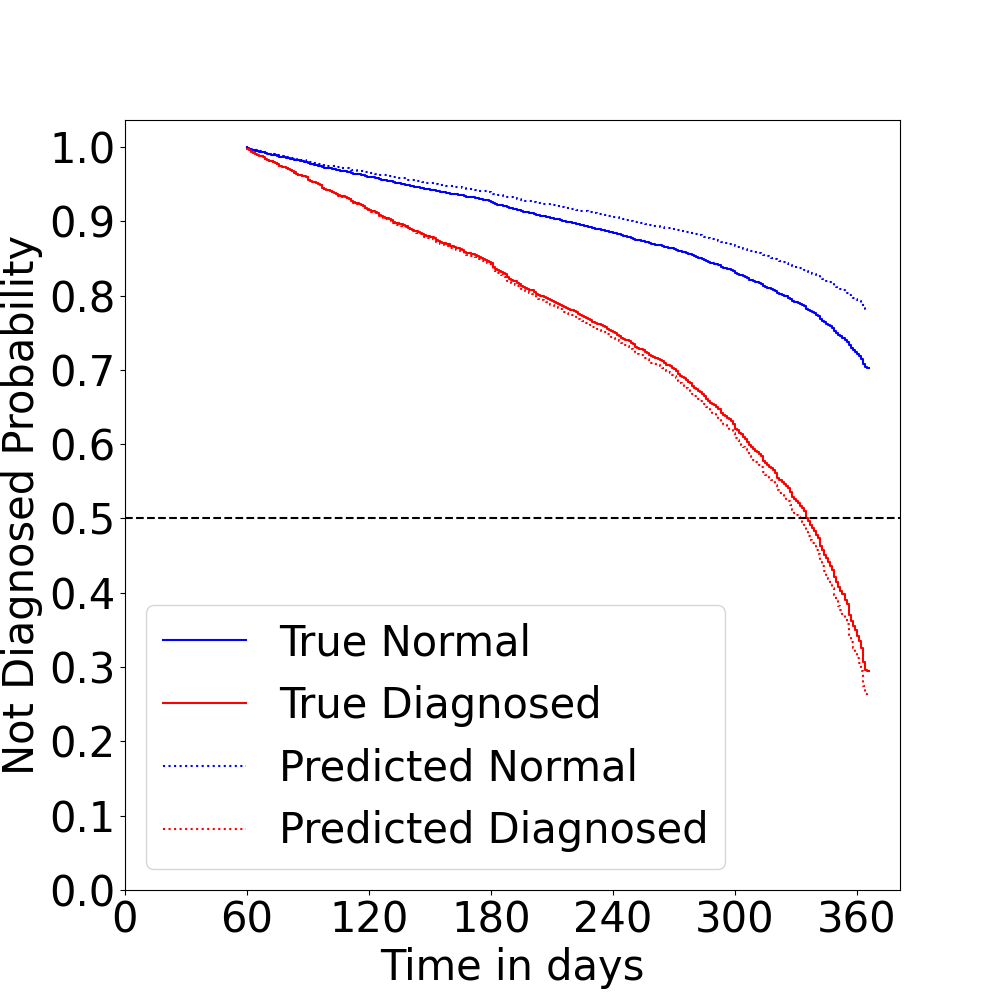}
	\caption{Approach 3: Distinct}
\end{subfigure}
	\caption{CKD}

\end{subfigure}
 \begin{subfigure}[b]{\textwidth}
	\begin{subfigure}[b]{0.32\textwidth}
		\includegraphics[width=\textwidth]{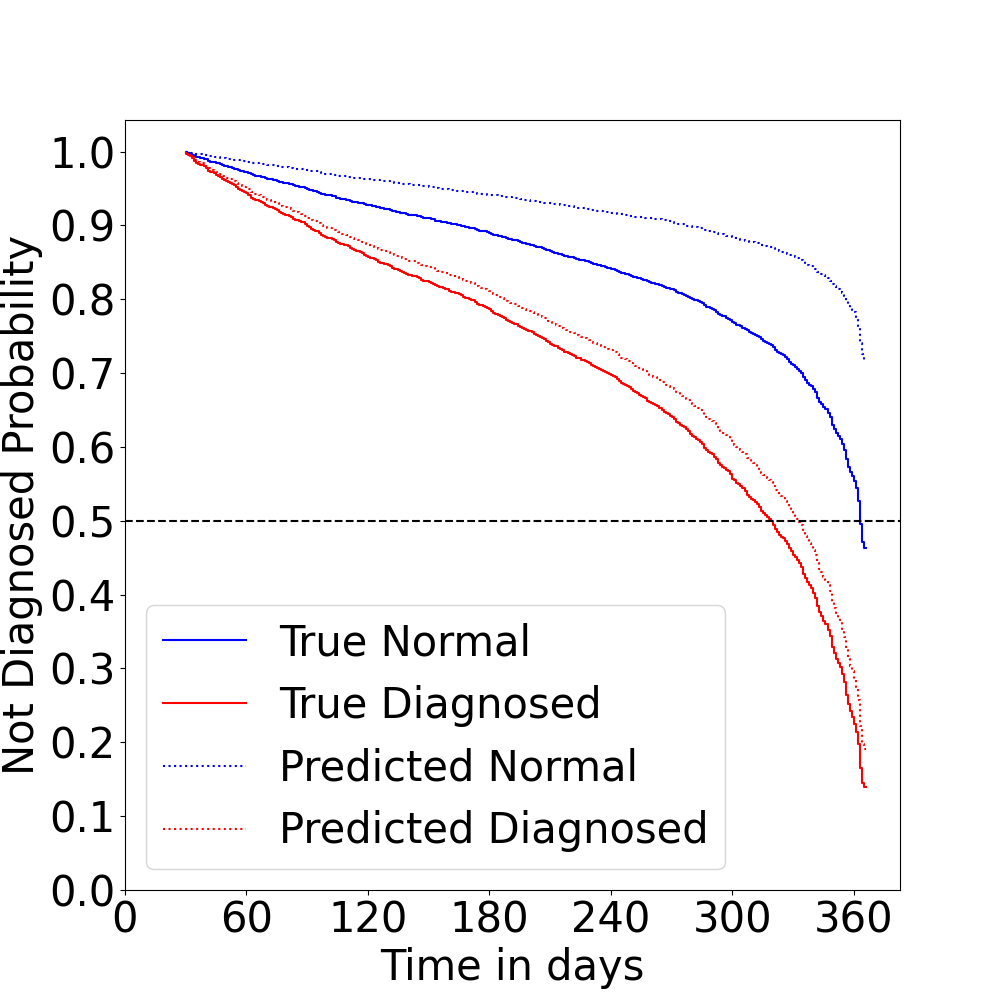}
		\caption{Approach 1: Similar}
	\end{subfigure}
\begin{subfigure}[b]{0.32\textwidth}
	\includegraphics[width=\textwidth]{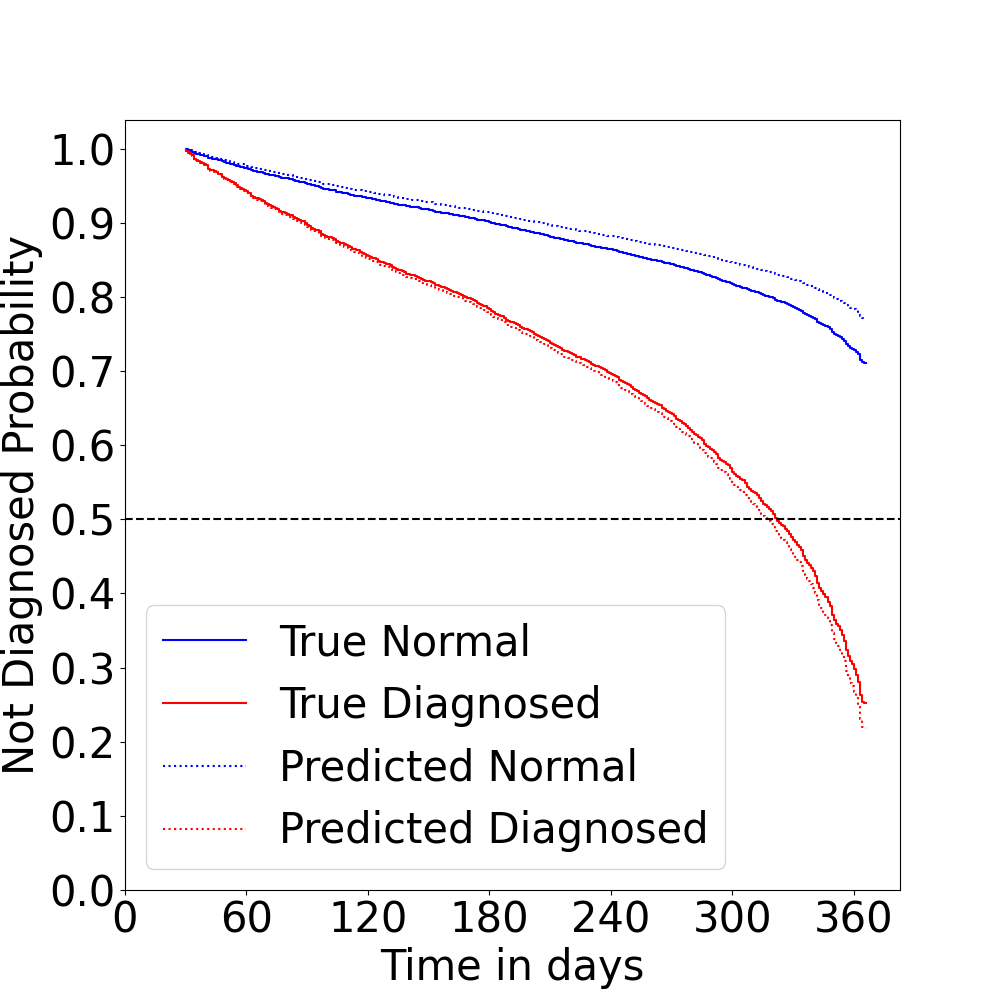}
	\caption{Approach 2: Overlap} 
\end{subfigure}
\begin{subfigure}[b]{0.32\textwidth}
	\includegraphics[width=\textwidth]{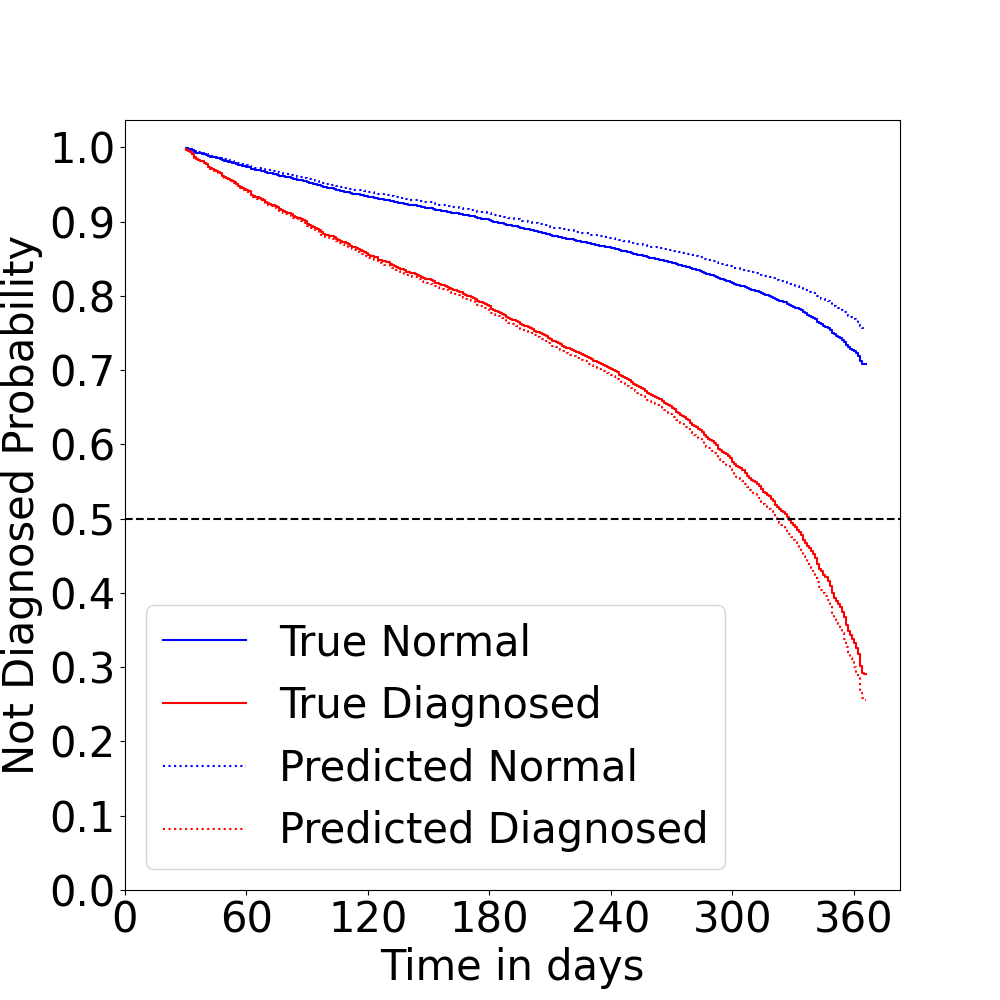}
	\caption{Approach 3: Distinct}
\end{subfigure}
	\caption{COPD}

\end{subfigure}
\end{figure}

\clearpage
\section{Explanations} \label{ap: expalanations}

\begin{figure}[ht!]
\vspace{-1cm}
	\centering
 \begin{subfigure}[b]{0.7\textwidth}
	\begin{subfigure}[b]{0.49\textwidth}
		\includegraphics[width=\textwidth]{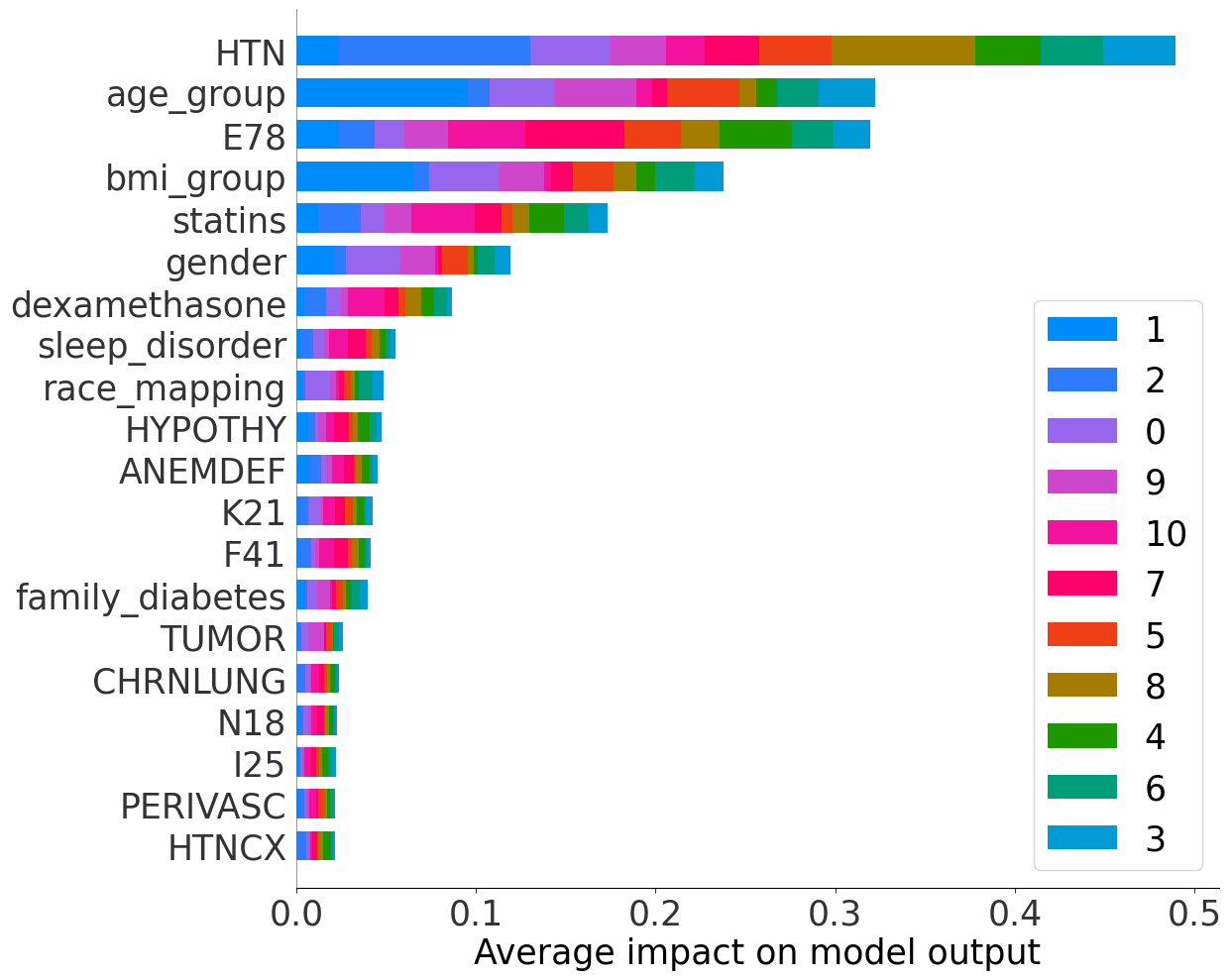}
		\caption{SurvSHAP}
	\end{subfigure}
\begin{subfigure}[b]{0.49\textwidth}
	\includegraphics[width=\textwidth]{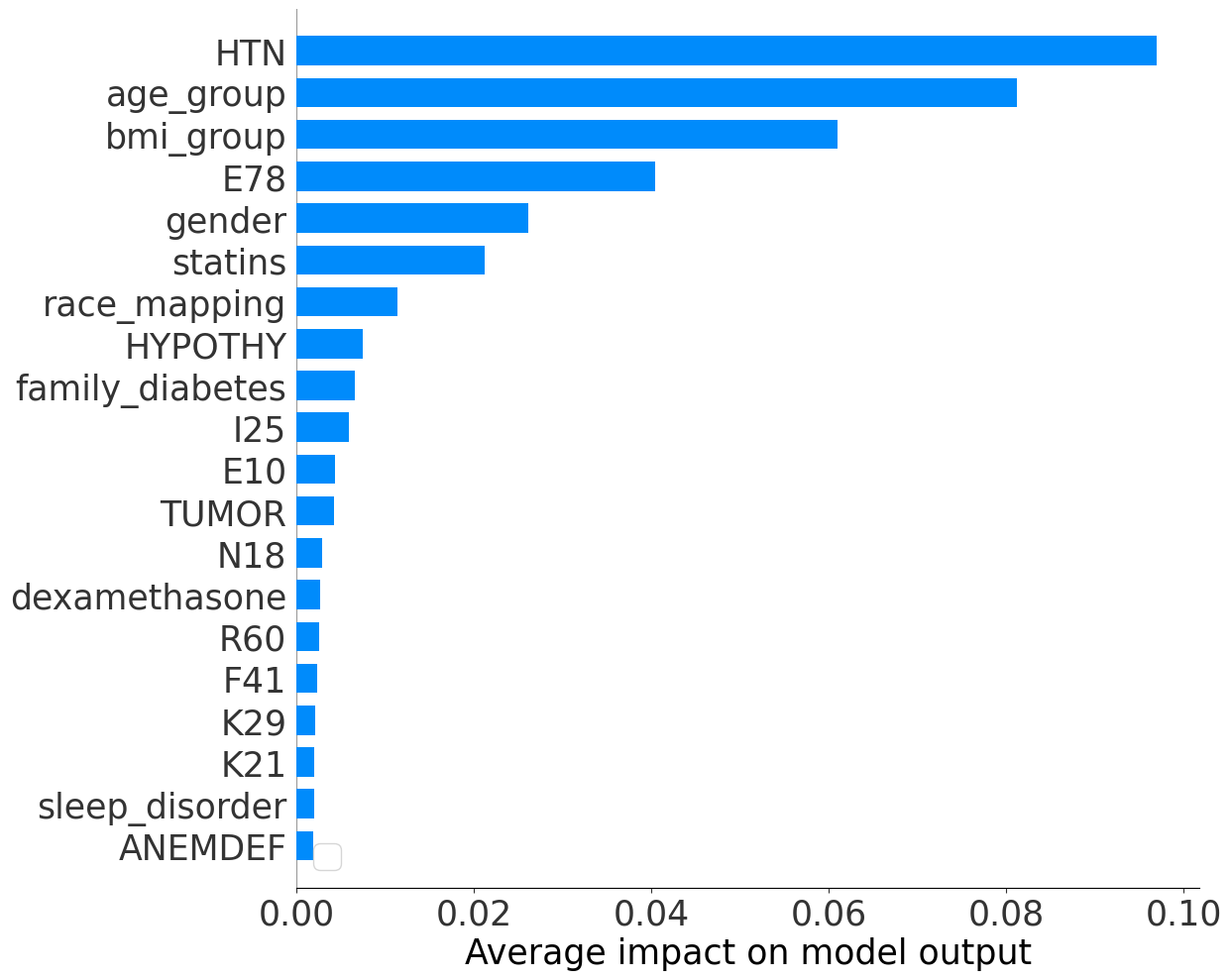}
	\caption{Custom } 
\end{subfigure}
	\caption{Diabetes}

\end{subfigure}
 \begin{subfigure}[b]{0.7\textwidth}
	\begin{subfigure}[b]{0.49\textwidth}
		\includegraphics[width=\textwidth]{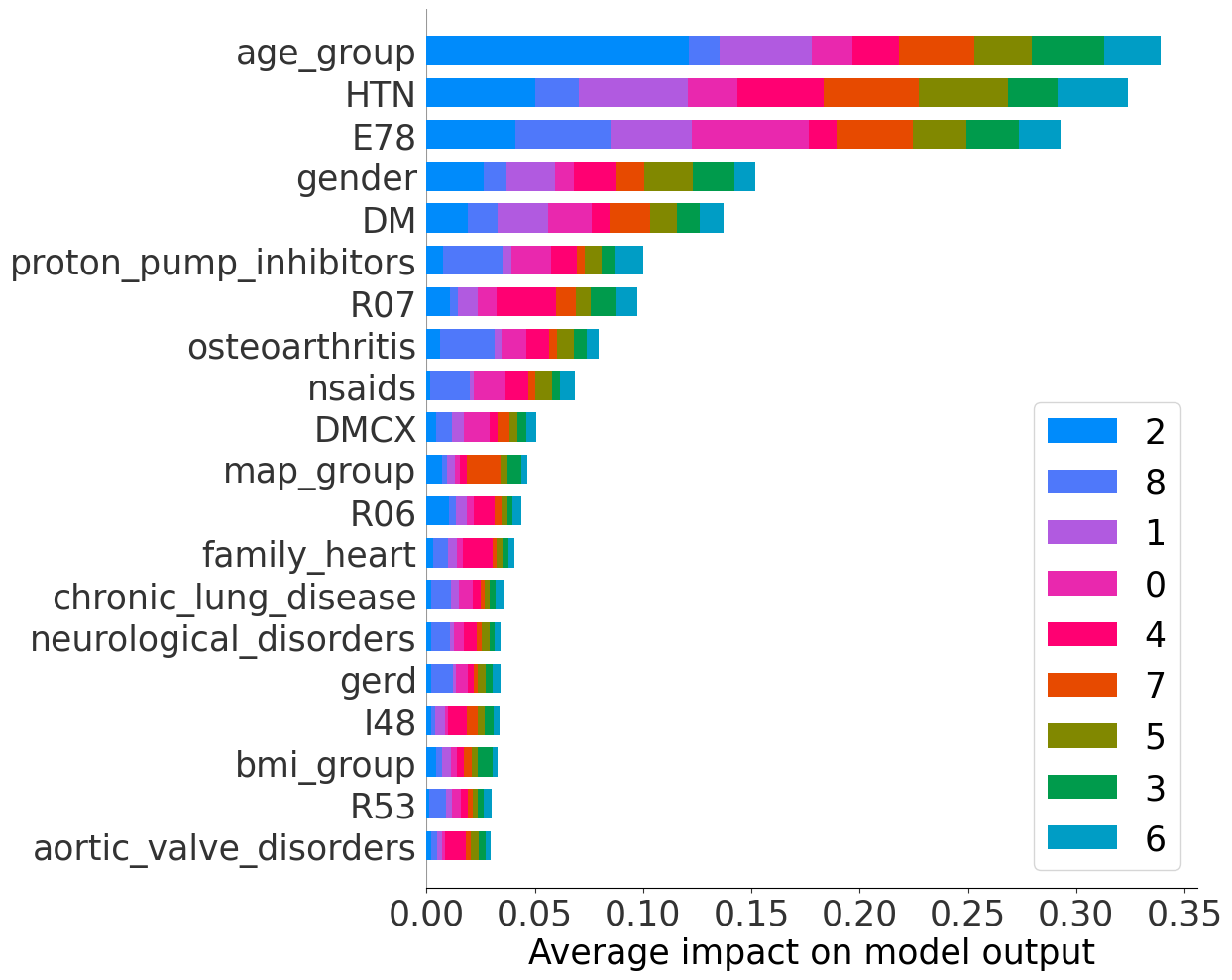}
		\caption{SurvSHAP}
	\end{subfigure}
\begin{subfigure}[b]{0.49\textwidth}
	\includegraphics[width=\textwidth]{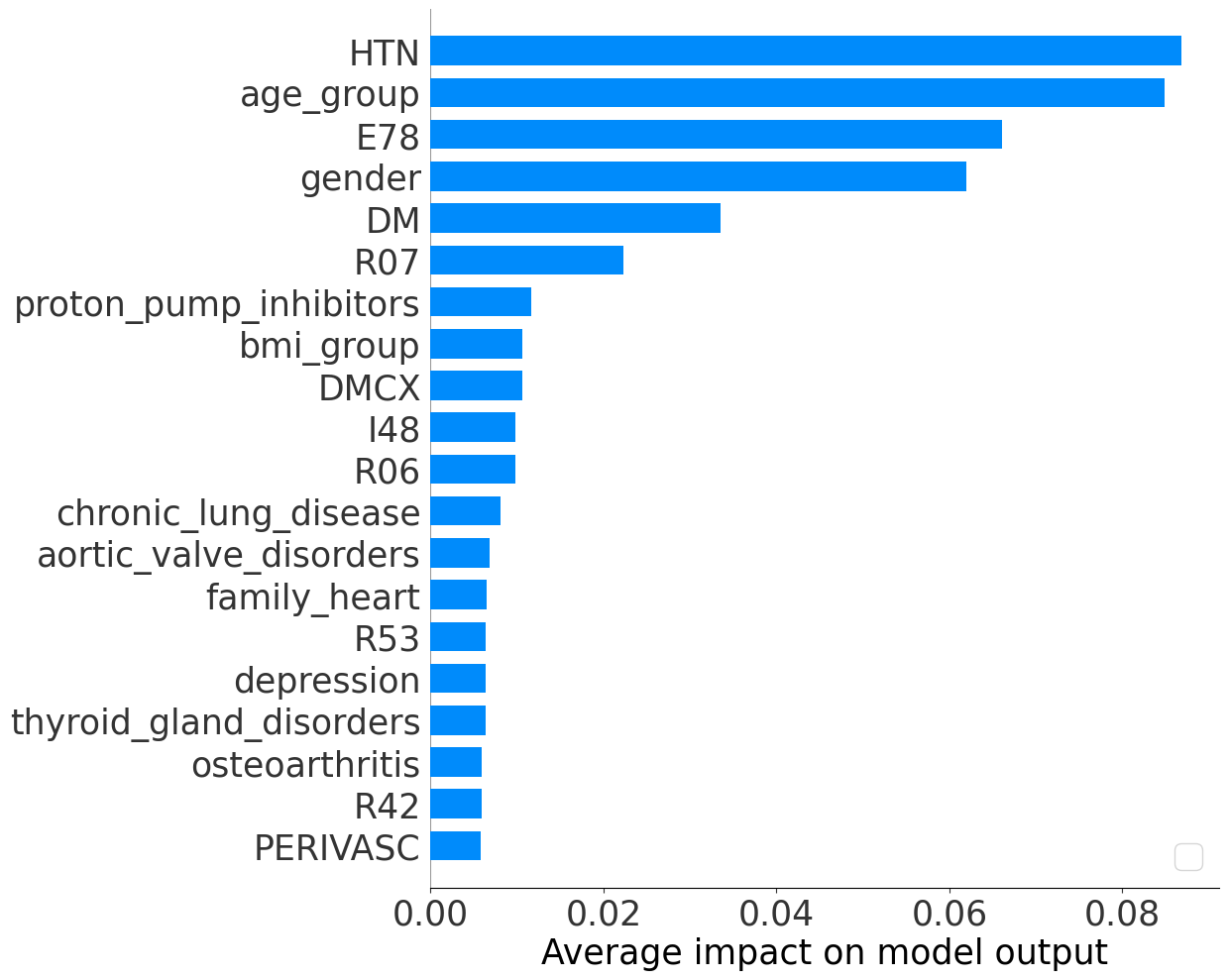}
	\caption{Custom } 
\end{subfigure}
	\caption{Heart}

\end{subfigure}
 \begin{subfigure}[b]{0.7\textwidth}
	\begin{subfigure}[b]{0.49\textwidth}
		\includegraphics[width=\textwidth]{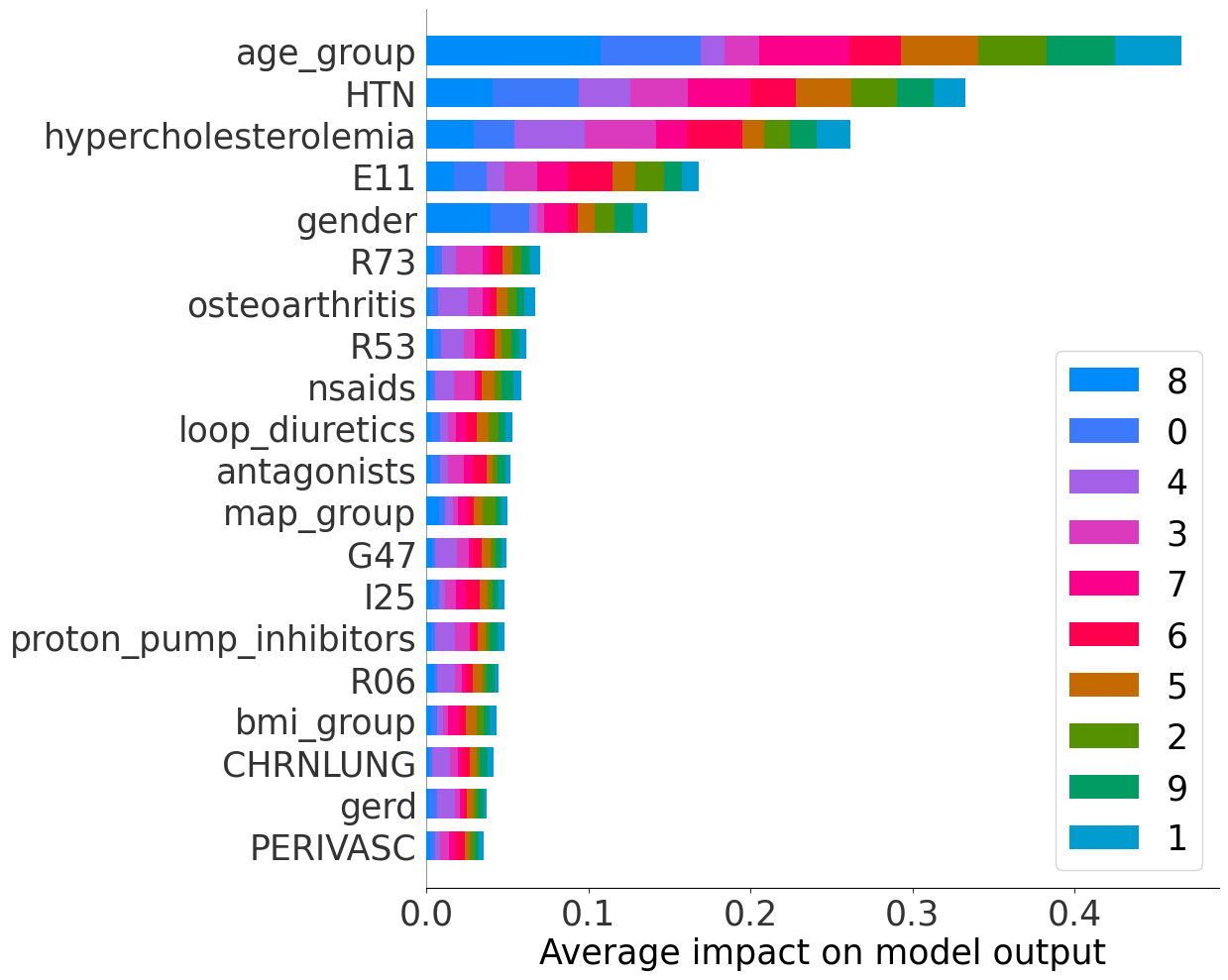}
		\caption{SurvSHAP}
	\end{subfigure}
\begin{subfigure}[b]{0.49\textwidth}
	\includegraphics[width=\textwidth]{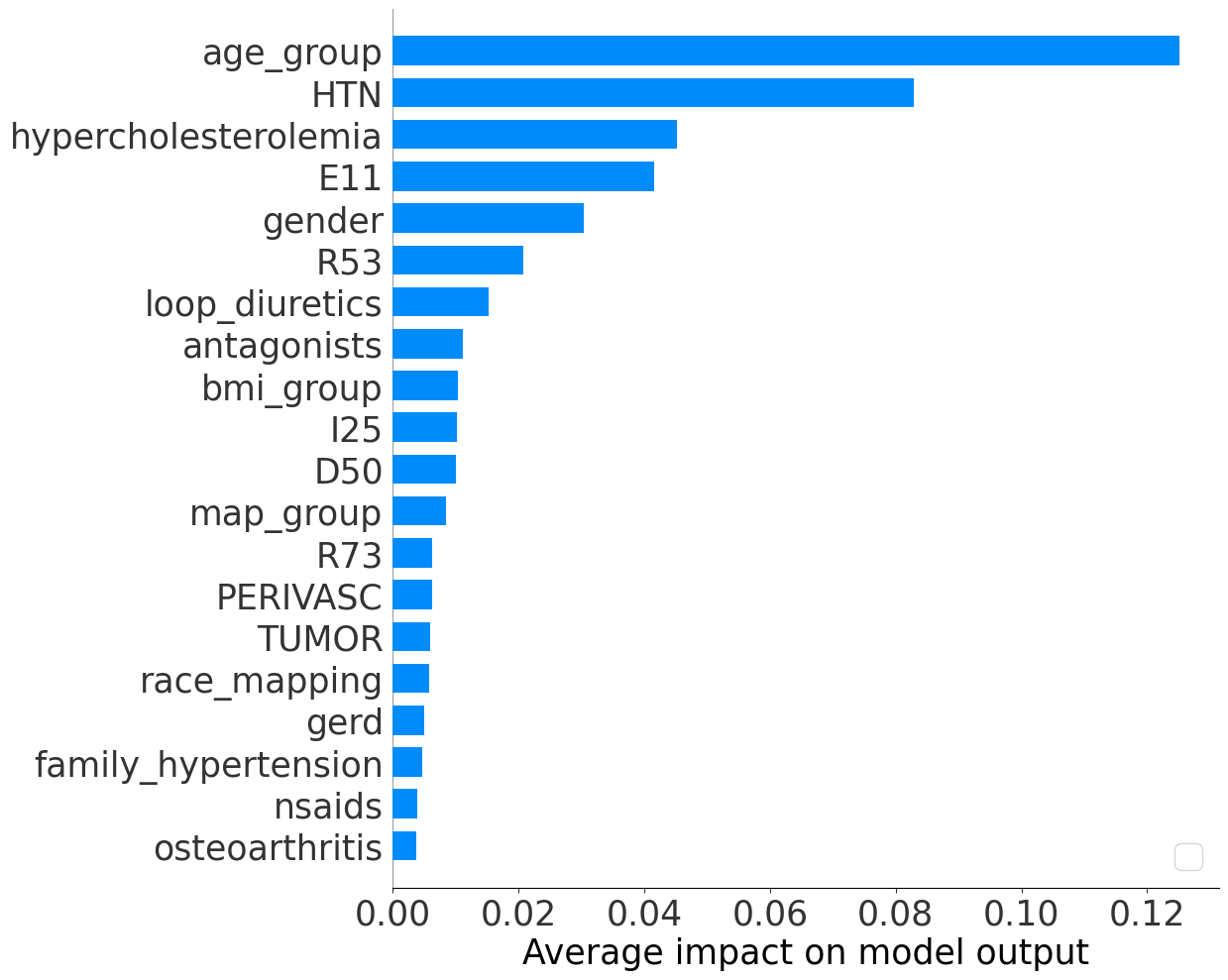}
	\caption{Custom} 
\end{subfigure}
	\caption{CKD}

\end{subfigure}
 \begin{subfigure}[b]{0.7\textwidth}
	\begin{subfigure}[b]{0.49\textwidth}
		\includegraphics[width=\textwidth]{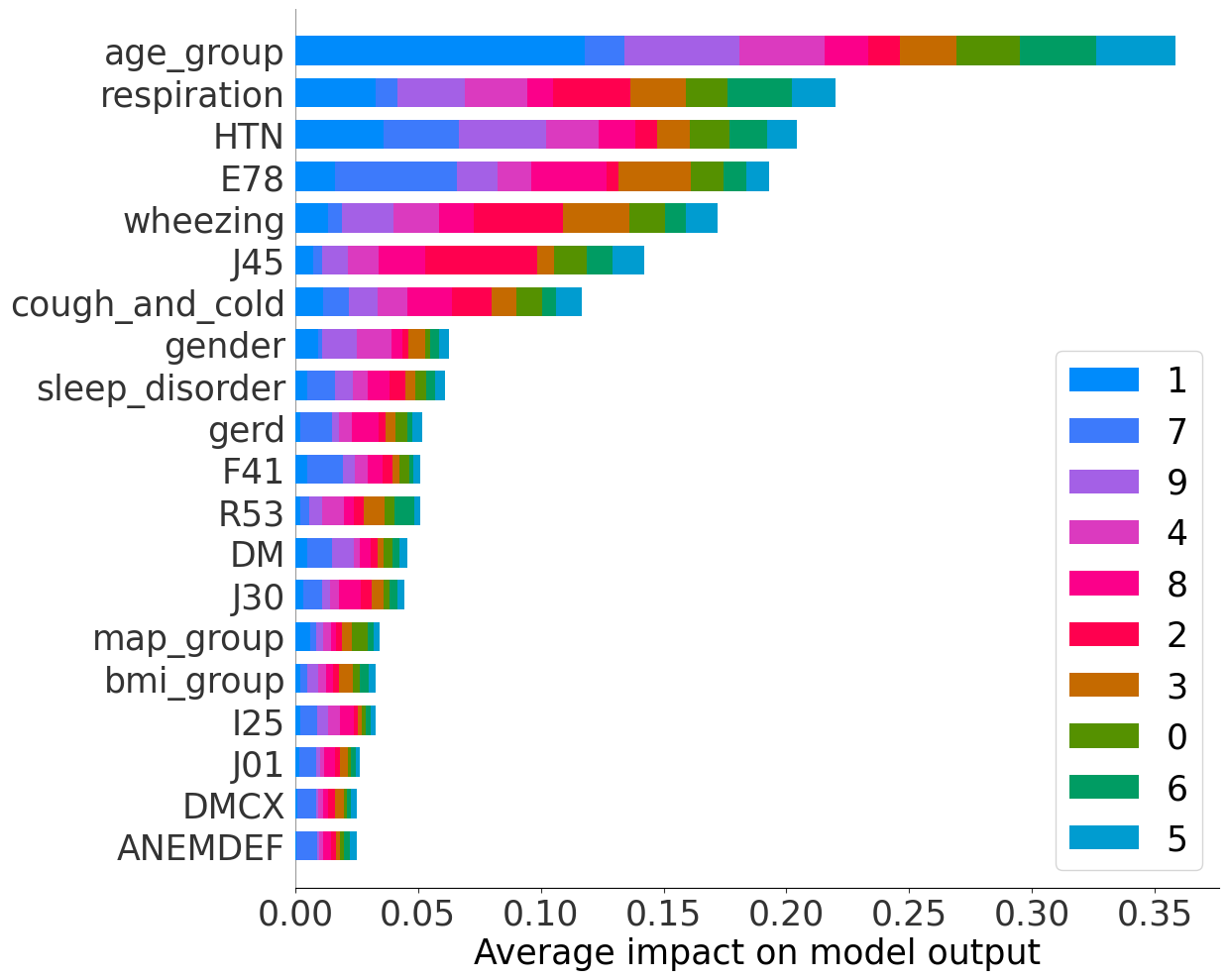}
		\caption{SurvSHAP}
	\end{subfigure}
\begin{subfigure}[b]{0.49\textwidth}
	\includegraphics[width=\textwidth]{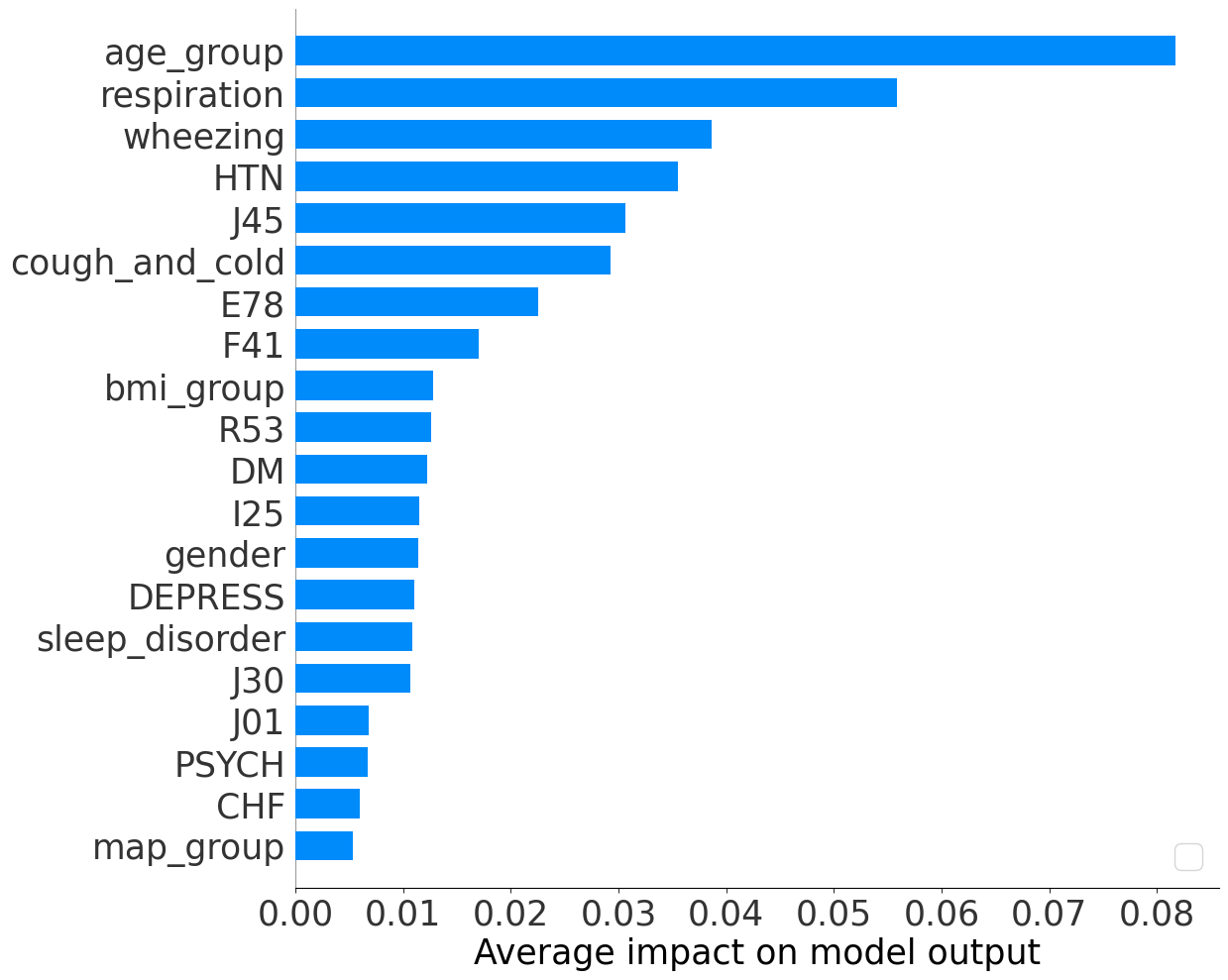}
	\caption{Custom} 
\end{subfigure}
	\caption{COPD}

\end{subfigure}
\end{figure}
\end{document}